\documentclass[10pt,twocolumn,letterpaper]{article}

\usepackage{cvpr}              
\usepackage{multirow}
\usepackage{makecell}
\definecolor{cvprblue}{rgb}{0.21,0.49,0.74}
\usepackage[pagebackref,breaklinks,colorlinks,allcolors=cvprblue]{hyperref}

\def\paperID{16048} 
\def\confName{CVPR}
\def\confYear{2025}

\title{Visual Consensus Prompting for Co-Salient Object Detection}

\author{Jie Wang\textsuperscript{1}, Nana Yu\textsuperscript{1}, Zihao Zhang\textsuperscript{1}, Yahong Han\textsuperscript{1}\thanks{Corresponding author.}\\
\textsuperscript{1}College of Intelligence and Computing, Tianjin University, Tianjin, China\\
{\tt\small wangjiexy@tju.edu.cn, yunana@tju.edu.cn, zhangzihao2490@tju.edu.cn,  yahong@tju.edu.cn}
}

\usepackage[ruled,vlined,linesnumbered]{algorithm2e}
\usepackage{colortbl}
\definecolor{customcolor}{HTML}{dbeef3} 

\begin{document}
\maketitle
\begin{abstract}
Existing co-salient object detection (CoSOD) methods generally employ a three-stage architecture (i.e., encoding, consensus extraction $\&$ dispersion, and prediction) along with a typical full fine-tuning paradigm. Although they yield certain benefits, they exhibit two notable limitations: 1) This architecture relies on encoded features to facilitate consensus extraction, but the meticulously extracted consensus does not provide timely guidance to the encoding stage. 2) This paradigm involves globally updating all parameters of the model, which is parameter-inefficient and hinders the effective representation of knowledge within the foundation model for this task. Therefore, in this paper, we propose an interaction-effective and parameter-efficient concise architecture for the CoSOD task, addressing two key limitations. It introduces, for the first time, a parameter-efficient prompt tuning paradigm and seamlessly embeds consensus into the prompts to formulate task-specific Visual Consensus Prompts (VCP). Our VCP aims to induce the frozen foundation model to perform better on CoSOD tasks by formulating task-specific visual consensus prompts with minimized tunable parameters. Concretely, the primary insight of the purposeful Consensus Prompt Generator (CPG) is to enforce limited tunable parameters to focus on co-salient representations and generate consensus prompts. The formulated Consensus Prompt Disperser (CPD) leverages consensus prompts to form task-specific visual consensus prompts, thereby arousing the powerful potential of pre-trained models in addressing CoSOD tasks. Extensive experiments demonstrate that our concise VCP outperforms 13 cutting-edge full fine-tuning models, achieving the new state of the art (with 6.8$\%$ improvement in $F_m$ metrics on the most challenging CoCA dataset). Source code has been available at \url{https://github.com/WJ-CV/VCP}.
\end{abstract}

\begin{figure}[!t]
	\centering
	\includegraphics[width=3.21in]{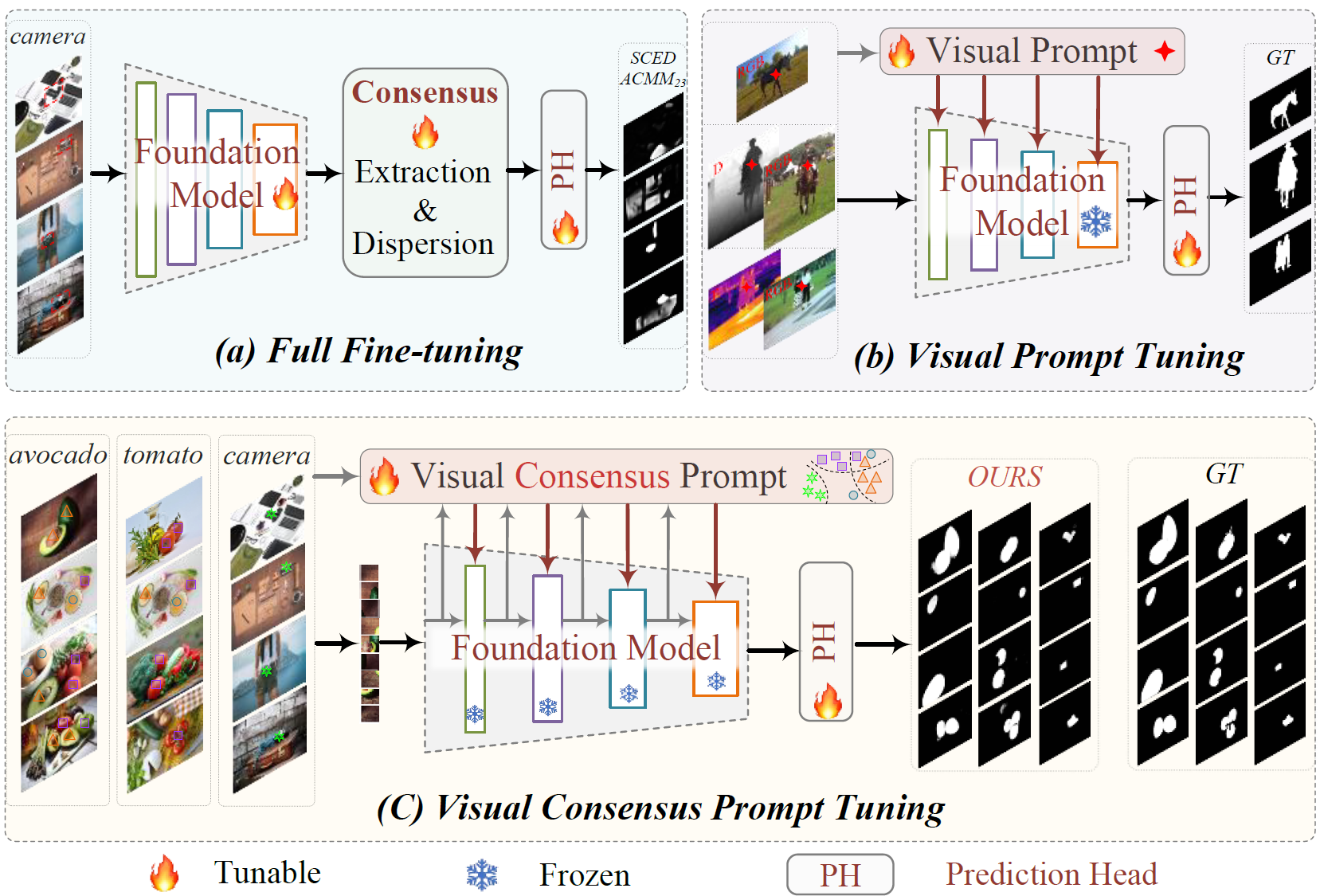}
	\vspace{-0.15in}
	\DeclareGraphicsExtensions.
	\begin{center}
		\caption{Existing relevant methods \textsl{VS.} our VCP. (a) Existing CoSOD methods based on typical architectural patterns and full fine-tuning paradigms. (b) Introducing simple tunable parameters as visual prompts to address foreground segmentation tasks in single-scene images. (c) Our proposed VCP and some visualization results. The frozen foundation model is mined to generate task-specific visual consensus prompts (with minimized tunable parameters), thereby inducing it to effectively perform CoSOD.}\label{fig-1}
	\end{center}
	\vspace{-0.43in}
\end{figure}    
\vspace{-13pt}
\section{Introduction}

\label{sec:intro}
Co-Salient Object Detection (CoSOD) is a group-based image understanding task aimed at detecting salient objects that commonly appear among a group of relevant images. CoSOD methods have demonstrated their effectiveness as a key pre-processing step for various computer vision tasks, such as co-segmentation \cite{zhang2022deep}, co-localization \cite{gong2020learning}, and object tracking \cite{yang2022co}. Additionally, with the effectiveness of CoSOD methods continue to improve, they are increasingly being utilized in numerous practical applications \cite{fan2021group, wang2021salient, zhang2016detection}. 

\begin{figure}[!t]
	\centering
	\includegraphics[width=3.05in]{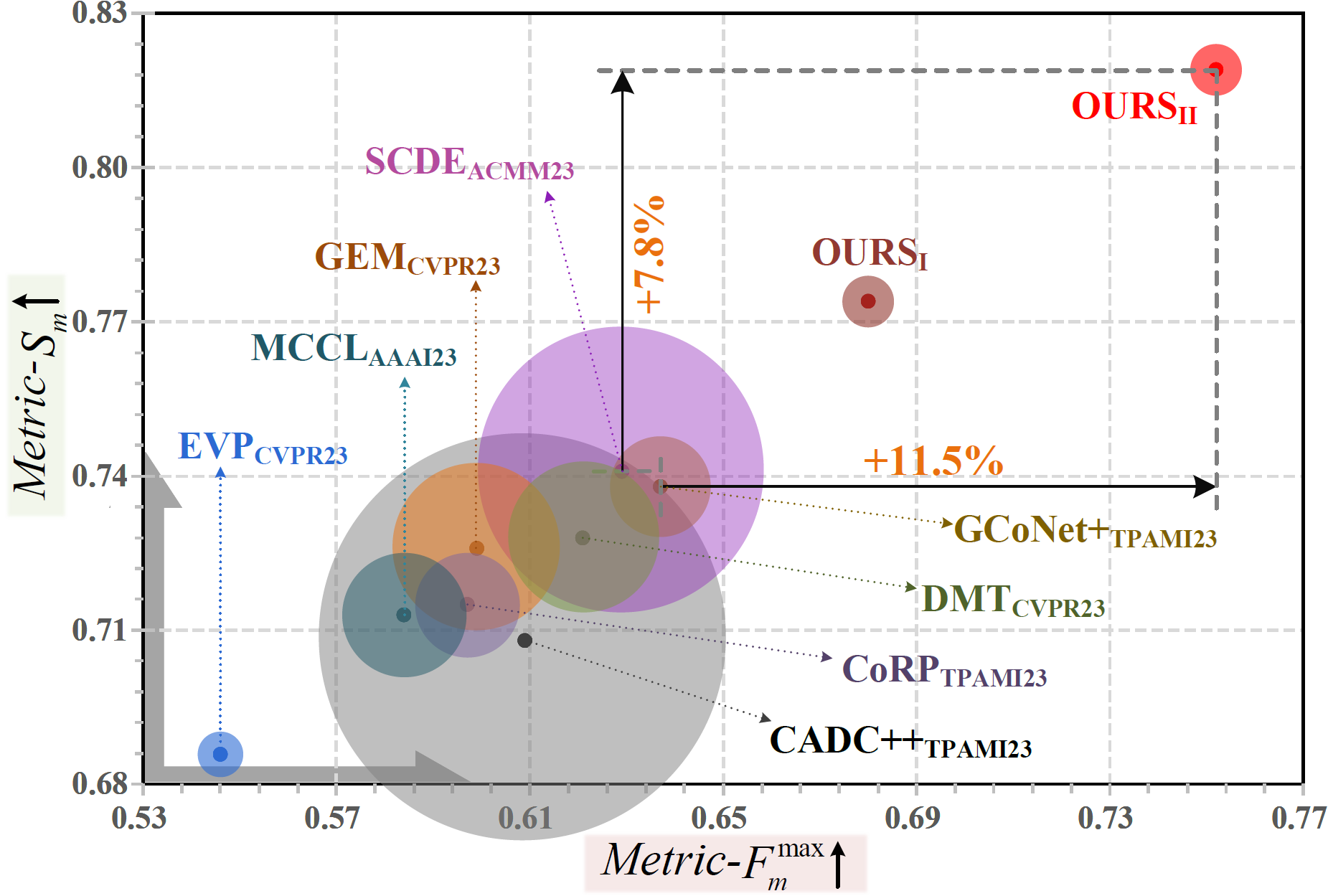}
	\vspace{-0.2in}
	\DeclareGraphicsExtensions.
	\begin{center}
		\caption{Quantitative comparison of our VCP with 8 representative methods on the CoCA \cite{zhang2020gradient} dataset regarding $S_m$, $F_m^{\max}$ metrics, and tunable parameters. The bubble area represents the tunable parameters (M). SCED \cite{xu2023co}, GEM \cite{wu2023co}, MCCL \cite{zheng2023memory}, GCoNet+ \cite{zheng2023gconet+}, CoPR \cite{zhu2023co}, DMT \cite{li2023discriminative}, and CADC++ \cite{zhang2023cadc++} are all full fine-tuning CoSOD methods. EVP \cite{liu2023explicit} is based on prompt learning for SOD tasks, and we retrain it using the CoSOD dataset.}\label{fig-2}
	\end{center}
	\vspace{-0.44in}
\end{figure}

Numerous impressive CoSOD methods have been proposed and continually advancing the performance of this task. As illustrated in Fig.\ref{fig-1} (a), most of these methods \cite{zhang2020gradient, zhang2021summarize, yu2022democracy, zheng2023memory, xu2023co, zhu2023co} employ a common three-stage architecture: first, encoding multi-scale features using a foundation model pretrained on large-scale datasets; next, designing unique consensus mining schemes to focus on and mine co-salient object representations within the group of relevant images to complete consensus extraction and further disperse the consensus into multi-scale features; finally, obtaining co-saliency predictions using a prediction head. In this architecture, the effectiveness of encoded features greatly facilitates the extraction and dispersion of consensus, but the meticulously extracted consensus struggles to guide feature encoding in a timely manner, as the encoder is only fine-tuned at the end of parameter optimization. In a nutshell, this architecture lacks efficient interaction between encoding and consensus, resulting in neither fully realizing their potential. More critically, existing methods typically employ the full fine-tuning paradigm, which involves adapting the model to the specific CoSOD task by tuning all parameters (including the large-scale pre-trained foundation model) using existing CoSOD datasets \cite{zhang2020gradient, lin2014microsoft, wang2019robust}. The full fine-tuning paradigm has two main limitations: 1) It is parameter-inefficient and requires many repetitions of tuning and storage of the entire pre-trained model, resulting in significant computational and storage overheads. 2) The quality and quantity of fine-tuning data also restrict the effectiveness of knowledge representation of the pre-trained foundation model in this task. These limitations make it difficult for existing methods to achieve more effective and efficient CoSOD performance (Fig.\ref{fig-1} (a) and Fig.\ref{fig-2}). Additionally, they increase the difficulty of CoSOD methods in practical applications, and this difficulty further increases with the inherent trend of increasing scale of foundation models (e.g., Transformer series or large models).

The parameter-efficient prompt tuning paradigm \cite{lester2021power, li2021prefix, liu2021p} has been proposed and promoted in Natural Language Processing (NLP) \cite{brown2020language, radford2019language, radford2018improving}. Inspired by that, recent works \cite{jia2022visual, nie2023pro, chen2022adaptformer, zhu2023visual} have introduced the prompt tuning paradigm into visual tasks, and guided the model to solve visual recognition tasks by freezing the foundation model and adding a few learnable parameters as task-specific visual prompts. Additionally, some promising works utilize the idea of prompt tuning to address foreground segmentation tasks (Fig.\ref{fig-1} (b)). These methods \cite{liu2023explicit, luo2024vscode} preset simple tunable embeddings as foreground prompts to address SOD-related tasks. As shown in Fig. \ref{fig-2}, these methods perform poorly when tackling specific CoSOD tasks, primarily due to two main limitations: (1) These methods solely focus on single-image foregrounds regardless of their category, while the CoSOD not only expands a group dimension but also involves numerous non-co-salient objects within the images as interference. (2) Introducing simple tunable parameters as visual prompts struggles to model intra-group co-salient representations and to achieve effective CoSOD performance. Hence, we ponder: How to formulate task-specific prompts for CoSOD to adapt the foundation model to effectively and efficiently facilitate this task?

At this juncture, it is surprising to find that if the extraction and dispersion of consensus (the key representations specific to the CoSOD task) are embedded into visual prompts, the architecture we seek for efficient interaction between consensus and encoding can perfectly align with the prompt tuning paradigm. In this spirit, we propose a concise and parameter-efficient Visual Consensus Prompting (VCP) for co-salient object detection. Our VCP aims to induce the frozen pre-trained model to effectively and efficiently execute CoSOD tasks by constructing task-specific visual consensus prompts (with minimized tunable parameters). Given the critical importance of visual consensus for CoSOD tasks, we design two key components to support the implementation of VCP: the Consensus Prompt Generator (CPG) and the Consensus Prompt Dispenser (CPD). The primary insight of CPG is to enforce limited tunable parameters to focus on intra-group co-salient representations from frozen embedding features, thereby generating consensus prompts. CPD leverages consensus prompts to form task-specific visual consensus prompts and adaptively harnesses the powerful potential of frozen Transformer layers in addressing CoSOD. We summarize the contribution of our work as follows:

\begin{itemize}
	\item An interaction-effective and parameter-efficient visual consensus prompts architecture for CoSOD is proposed, which can serve as a powerful alternative to existing methods based on the common architecture and the full fine-tuning paradigm.
	
	\item The proposed Consensus Prompt Generator systematically aggregates task-specific consensus prompts, while the Consensus Prompt Disperser adaptively induces the foundation model to perform better in this task.
	
	\item Extensive experiments demonstrate the comprehensive performance advantage of our VCP. Particularly on the COCA dataset, which best reflects the model's robustness, our VCP outperforms the state-of-the-art method by \textbf{5.6$\%$} and \textbf{6.8$\%$} in terms of $S_m$ and $F_m$, respectively.
\end{itemize}

\begin{figure*}[!t]
	\centering
	\includegraphics[width=6.8in]{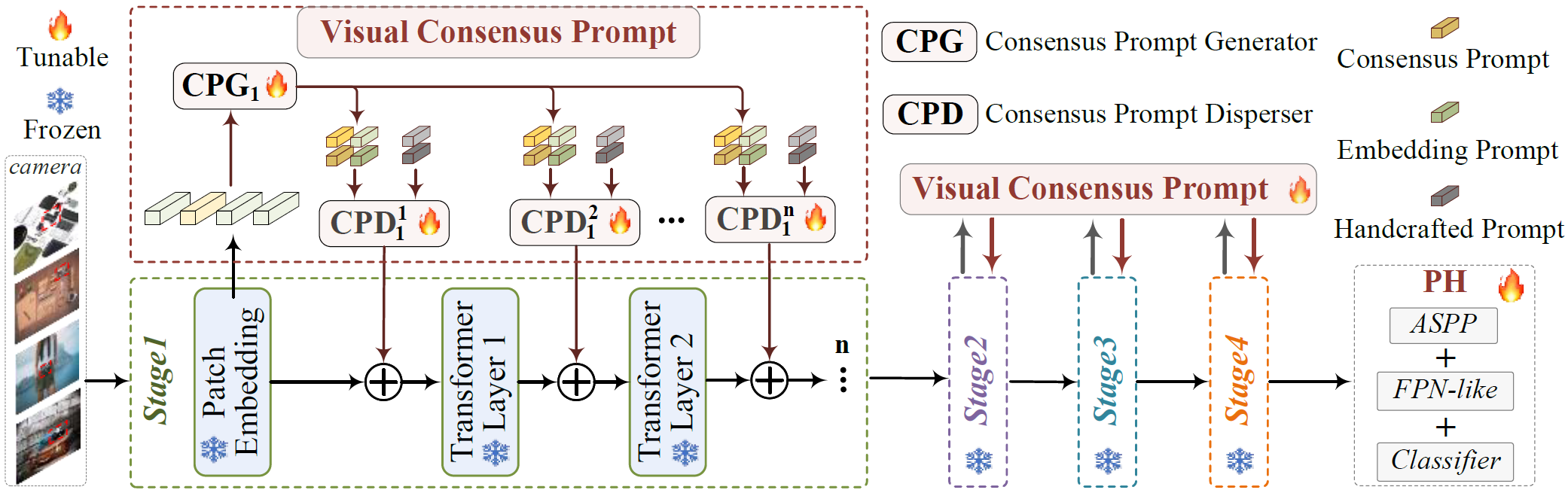}
	\DeclareGraphicsExtensions.
	\begin{center}
		\vspace{-0.111in}
		\caption{Overall framework pipeline of our proposed concise and parameter-efficient VCP model. We induce the frozen foundation model to perform better on the CoSOD task by formulating Visual Consensus Prompts with minimal tunable parameters. The proposed Consensus Prompt Generator (CPG) and Consensus Prompt Disperser (CPD) support the implementation of VCP. The CPG mines intra-group co-salient representations of the frozen embeddings to generate consensus prompts ${P_{Co}}$. The CPD utilizes ${P_{Co}}$ to form Visual Consensus Prompts and induce the frozen transformer layers to perform the CoSOD task.}\label{fig-3}
	\end{center}
	\vspace{-0.365in}
\end{figure*}
It’s our belief that the CoSOD task can be conducted with a concise architecture in an interaction-effective and parameter-efficient manner. We hope the impressive performance of VCP can demonstrate the potential of the consensus prompt tuning paradigm to the CoSOD community.

\section{Related works}

\subsection{Co-Salient Object Detection}
CoSOD aims to discover and segment salient objects that appear commonly within a group of images. Recent CoSOD methods \cite{fan2021group, zhang2020gradient, zhang2021summarize, yu2022democracy, zheng2023memory, xu2023co, wang2024single} have achieved impressive performance, and their implementation paradigms can be broadly summarized as consensus extraction and consensus dispersion. Consensus extraction aims to focus on the salient objects common to the group of images and form a consensus representation. Consensus dispersion utilizes the extracted consensus to guide the model's decoding and generate predictions.  Xu \textit{et al.} \cite{xu2023co} employ a hierarchical Transformer to extract semantic consensus and propagate consensus based on the Transformer. Some promising methods have attempted to use saliency predictions to filter out non-salient backgrounds or to further refine consensus representations \cite{zheng2023gconet+, jin2020icnet, zhu2023co}. However, existing methods lack effective interaction between encoding and consensus, and they are all based on the full fine-tuning strategy, which involves adapting the model to the specific CoSOD task by adjusting all parameters (including the large-scale pre-trained foundation model) using existing CoSOD datasets. This parameter-inefficient tuning paradigm incurs significant computational and storage overheads, making it not only challenging to achieve better performance but also unfriendly to practical applications. In contrast, an interaction-effective and parameter-efficient architecture for CoSOD is proposed, which can serve as a powerful alternative to existing prevailing methods.
   
\subsection{Visual Prompting Tuning}
Prompt-based learning is pioneered and driven by the GPT series \cite{brown2020language, radford2019language, radford2018improving} in the field of NLP. It initially referred to adding task-specific descriptions to downstream inputs to assist language models in handling downstream tasks, rather than solely adapting pre-trained models to fit downstream tasks through full fine-tuning. In addition to subsequent work on how to construct better prompt texts \cite{jiang2020can, shin2020autoprompt}, recent efforts have also proposed treating prompts as continuous vectors specific to tasks and directly optimizing them through gradients during the fine-tuning process, known as Prompt Tuning \cite{lester2021power, li2021prefix, liu2021p}. Besides, prompt learning also shows its effectiveness in many computer vision tasks \cite{liu2023explicit, zhu2023visual, luo2024vscode, jia2022visual, nie2023pro, chen2022adaptformer}. VPT \cite{jia2022visual} provides a set of learnable parameters pre-prepared for Transformer encoders, significantly outperforming full fine-tuning on 20 downstream identification tasks. Unlike simply adding some linear layers or tunable parameters to address visual recognition tasks \cite{jia2022visual, nie2023pro, chen2022adaptformer}, some promising methods incorporate prompt learning into foreground segmentation tasks. EVP \cite{liu2023explicit} introduces handcrafted features to create explicit visual prompts for effective single-image SOD. VSCode \cite{luo2024vscode} incorporates task-specific and domain-specific prompts to address single-modal or multi-modal SOD tasks through full fine-tuning of the model. These methods primarily focus on foreground objects in a single scene and only perform simple prompt tuning, making it difficult to model intra-group consensus and achieve satisfactory co-salient object segmentation results.

\section{Proposed Methodology}
\subsection{Overview}

\begin{figure}[!t]
	\centering
	\includegraphics[width=3.25in]{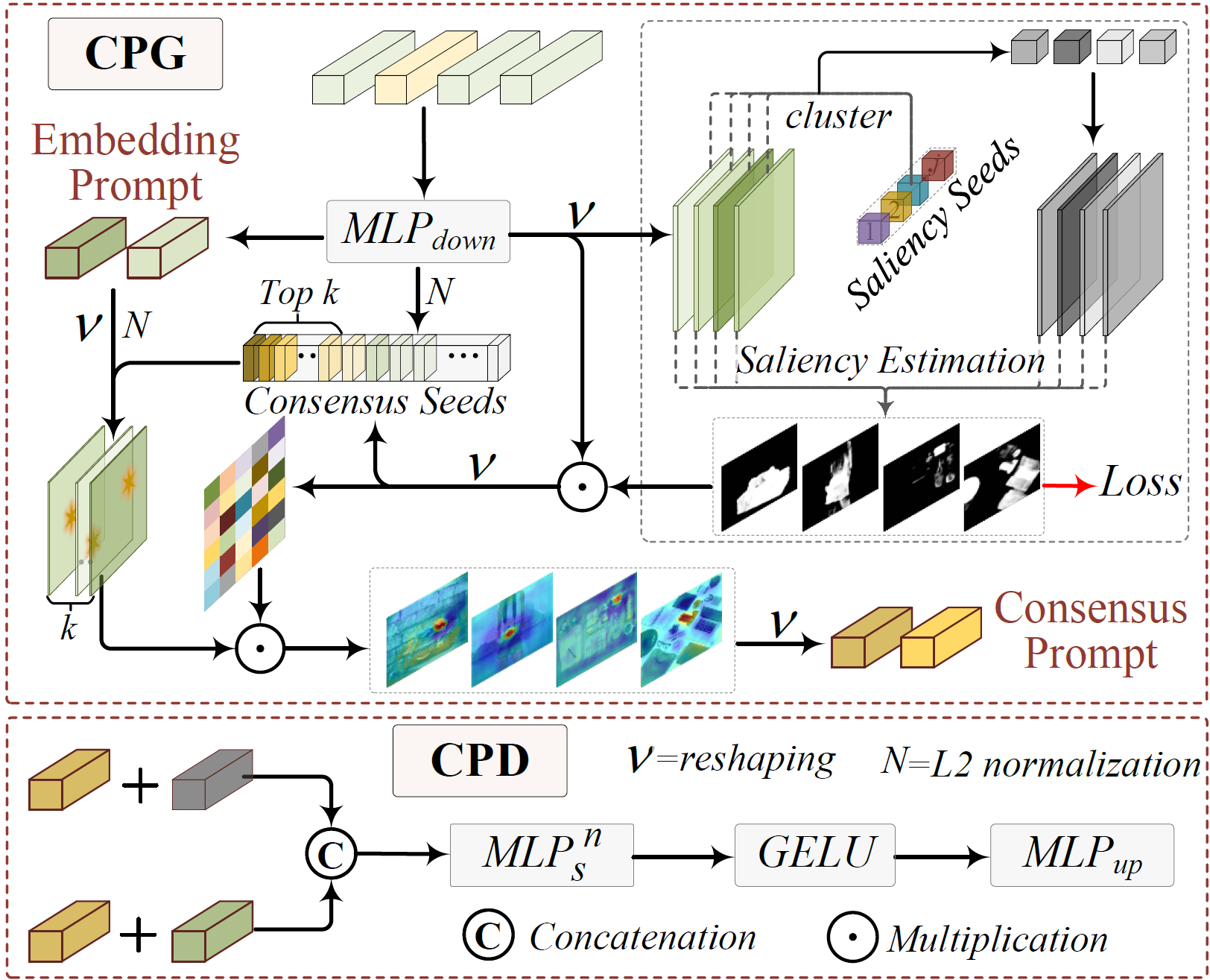}
	\DeclareGraphicsExtensions.
	\begin{center}
		\vspace{-0.15in}
		\caption{Overall pipeline of the proposed CPG and CPD. The CPG utilizes predefined saliency seeds to generate saliency estimation maps through clustering, thereby obtaining consensus seeds. By selecting top-k representative consensus seeds, consensus prompts ${P_{Co}}$ are obtained. The CPD utilizes ${P_{Co}}$ to generate visual consensus prompts $P_{Visual}^{Co}$ and induce the frozen transformer layers to address the CoSOD task.}\label{fig-4}
	\end{center}
	\vspace{-0.4in}
\end{figure}

In this section, we propose Visual Consensus Prompt tuning, i.e., VCP, for CoSOD, it is a concise and parameter-efficient alternative to existing full fine-tuning methods \cite{xu2023co, zhu2023co, wu2023co, li2023discriminative, zhang2023cadc++}. The proposed VCP is a predefined prompt specific to the CoSOD task, which reformulates this task to closely resemble those addressed during the original pre-training, thereby inducing the foundation model to perform well on this task. Concisely, the core of our VCP is tuning only a few task-specific parameters to adapt large-scale pre-trained Transformer models for effectively addressing CoSOD tasks. For CoSOD tasks, given a set of related images $\{ {I_i} \in {\mathbb{R}^{C \times H \times W}}\} _{i = 1}^N$, the model is required to search for and segment intra-group co-salient objects in order to generate co-salient prediction maps $\{ {M_i}\} _{i = 1}^N$. The main insight of VCP is to enforce limited tunable parameters to focus on and generate task-specific consensus prompts, and utilize these prompts to induce the foundation model to perform effectively. As shown in Fig. \ref{fig-3}, the proposed VCP primarily consists of two key components: the CPG and the CPD. The CPG is responsible for mining intra-group co-salient representations of the frozen embeddings to generate consensus prompts. The CPD utilizes these consensus prompts to form  task-specific visual consensus prompts and induce the frozen transformer layers to perform the CoSOD task.

Segformer \cite{xie2021segformer} is employed as our foundation model, which is a hierarchical Transformer architecture pretrained on ImageNet, utilized for semantic segmentation tasks. Segformer extracts multi-scale features through four stages, each comprising an overlapping patch embedding layer and several visual transformer blocks. The extracted embedding features $E = \{ {E_s} \in {\mathbb{R}^{N \times {L_s} \times {C_s}}},{L_s} = {H_s}{W_s}\} _{s = 1}^4$ are utilized by CPG to mine intra-group co-salient representations for generating consensus prompts ${P_{Co}}$ and embedding prompts ${P_{Em}}$. Inspired by traditional methods \cite{huang2008detection, popescu2005exposing} that employ handcrafted image features to aid segmentation tasks, we utilize fast Fourier transform \cite{liu2023explicit}, which generates Handcrafted Prompt ${P_{Hand}}$ through several predefined tunable embedding layers. CPD leverages consensus prompts to integrate three types of prompts (${P_{Co}}$, ${P_{Em}}$ and ${P_{Hand}}$) to form task-specific visual consensus prompts $P_{Visual}^{Co}$, thereby adaptively inducing different depths of transformer layers. We modify the original decoding part of Segformer, reconstructing a concise Prediction Head (PH) and integrating a classifier.

\subsection{Consensus Prompt Generator}
Effectively mining intra-group consensus representations and suppressing irrelevant feature components is crucial for specific CoSOD tasks. Therefore, we propose a CPG, which can efficiently extract co-salient representations from frozen embedding features within the group and generate consensus prompts ${P_{Co}}$. As depicted in Fig. \ref{fig-4}, initially, the CPG initializes $j$ learnable saliency seed tensors, which are utilized to generate prototype representations of intra-image salient objects via clustering. These prototype representations guide the model in performing the initial saliency estimation. Subsequently, leveraging the obtained saliency estimation map, we generate pixel embeddings for all salient objects, i.e., consensus seeds. From these consensus seeds, $k$ highly relevant pixel embeddings are selected to form consensus representations, which are then mapped back to the original embeddings to generate consensus prompts ${P_{Co}}$.

The embedding features obtained by the foundation model have a high dimensionality. A tunable linear layer (with a scale $r$ to control the tunable parameters) is used to reduce the dimensionality of the input embedding features to form the initial embedding prompt ${P_{Em}} \in {\mathbb{R}^{N \times {L_s} \times {C_r}}},{C_r} = {C_s}/r$. Inspired by \cite{fan2021re, jin2020icnet, zhu2023co}, we leverage saliency priors to assist in consensus extraction. However, unlike previous approaches, we no longer rely on additional large-scale saliency detection datasets to aid in training the prediction head. Instead, we leverage the concept of prototype learning by learning saliency seeds (cluster centers) within the embedding to accomplish salient object representation. Specifically, $j$ learnable saliency seeds ${S_{seed}} \in {\mathbb{R}^{j \times {C_r}}}$ are pre-defined. The reshaped embedding features ${P_{em}} \in {\mathbb{R}^{N \times {C_r} \times {H_s} \times {W_s}}}$ are first subjected to normalization and convolution operations to form soft allocation probability scores ${S_{soft}} \in {\mathbb{R}^{N \times j \times {L_s}}}$.
\vspace{-0.05in}
\begin{equation}
\begin{array}{c}
{S_{soft}} = \nu (Softmax(conv({L_2}({P_{em}})))),
\end{array}
\end{equation}
where ${L_2}$ and $\nu $ represent L2 normalization and reshaping operations, respectively. Then, based on the ${S_{soft}} \in {\mathbb{R}^{N \times j \times {L_s}}}$, the residuals $Res \in {\mathbb{R}^{N \times j \times {C_r} \times {L_s}}}$ between the embedding features and the saliency seeds ${S_{seed}} \in {\mathbb{R}^{j \times {C_r}}}$ are computed. 
\vspace{-0.05in}
\begin{equation}
\begin{array}{c}
Res = (\nu ({P_{em}}) - \nu ({S_{seed}})) \times \nu ({S_{assign}}),
\end{array}
\end{equation}
These residuals $Res \in {\mathbb{R}^{N \times j \times {C_r} \times {L_s}}}$ are then utilized to weightedly sum the embedding features, resulting in updated saliency seeds representations $S_{seed}^{update} \in {\mathbb{R}^{N \times j \times {C_r}}}$. 
\begin{equation}
\begin{array}{c}
S_{seed}^{update} = \sum\limits_{i = 1}^{{L_s}} {Res} (N,j,{C_r},i),
\end{array}
\end{equation}
Finally, the updated saliency seeds are used to enhance the salient objects within the embedding features, thereby generating the saliency estimation map $\{ {M^s}\} _{s = 1}^4$.
\begin{equation}
\begin{array}{c}
{M^s} = conv[MLP({L_2}(S_{seed}^{update})),{P_{em}}].
\end{array}
\end{equation}
Where $[\cdot]$ represents channel concatenation. We utilize a prototype learning-based approach to achieve attention on salient objects in complex scenes by learning saliency prototypes present in clustered embeddings. Simultaneously, we utilize CoSOD labels for multi-stage supervision to maintain consistency between the saliency estimation maps and the consensus attention targets. We utilize the obtained saliency estimation map to further filter out non-co-salient object representations in the embedding features, forming consensus prompts ${P_{Co}} \in {\mathbb{R}^{N \times {L_s} \times {C_r}}}$. Specifically, we utilize the saliency estimation map $\{ {M^s}\} _{s = 1}^4$ to reshape all embedding representations ${P_{em}} \in {\mathbb{R}^{N \times {C_r} \times {H_s} \times {W_s}}}$ within the group into pixel patch embeddings, namely, consensus seeds $C{o_{seed}} \in {\mathbb{R}^{N{L_s} \times {C_r}}}$. Next, we search within the consensus seeds for the top-k pixel embeddings as representative consensus seeds $Co_{seed}^{rep} \in {\mathbb{R}^{k \times {C_r}}}$, which exhibit the highest correlation among them.
\vspace{-0.05in}
\begin{equation}
\begin{array}{c}
score = C{o_{seed}} \otimes \nu {(average({P_{em}} \times {M^s}))^T}, 
\end{array}
\end{equation}
\vspace{-0.2in}
\begin{equation}
\begin{array}{c}
Co_{seed}^{rep} = gather(C{o_{seed}},argtopk(score)), 
\end{array}
\end{equation}
where $\otimes$ represents matrix multiplication. Finally, the obtained representative consensus seeds are mapped back to the original embeddings to form effective consensus features, and spatial attention ${S_{att}} \in {\mathbb{R}^{N \times 1 \times {H_s} \times {W_s}}}$ is utilized to further enhance the consensus representation. The enhanced consensus features are reshaped to form consensus prompts ${P_{Co}} \in {\mathbb{R}^{N \times {L_s} \times {C_r}}}$.
\vspace{-0.05in}
\begin{equation}
\begin{array}{c}
{F_{Co}} = conv({L_2}({P_{em}}),weight = \nu (Co_{seed}^{rep})),
\end{array}
\end{equation}
\vspace{-0.2in}
\begin{equation}
\begin{array}{c}
{P_{Co}} = \nu (conv({F_{Co}} \times {S_{att}})).
\end{array}
\end{equation}
\vspace{-0.1in}


\begin{table*}[!t]
	\centering
	\normalsize
	\tabcolsep=0.001cm
	\renewcommand{\arraystretch}{0.99}
	\caption{
		Quantitative comparison between the proposed VCP and 13 SOTA methods on three benchmark datasets regarding six comprehensive quantitative metrics, tunable parameters, and model size. DUT-class \cite{zhang2020gradient}, COCO-9k \cite{lin2014microsoft}, and COCO-SEG \cite{wang2019robust} are wildly used training datasets in CoSOD and we denote them as D, C and S, respectively. “↑” means that the higher the numerical value, the better the model performance. Red, blue, and green represent the top three performances, respectively.} \label{tab-1}
	\vspace{-0.1in}
	\scalebox{0.77}{
		\begin{tabular}{cc|c|ccccccccccccccc}
			\toprule
			\multicolumn{1}{c|}{\multirow{3}[2]{*}{Datasets}} & \multicolumn{1}{c|}{\multirow{3}[2]{*}{Metrics}} & \multicolumn{1}{c|}{EVP} &   \multicolumn{1}{c}{CADC} & \multicolumn{1}{c}{GCoNet} & \multicolumn{1}{c}{DCFM} & \multicolumn{1}{c}{DMT} & \multicolumn{1}{c}{CoRP} & CADC++ & \multicolumn{1}{c}{GEM} & \multicolumn{1}{c}{GCoNet+} & \multicolumn{1}{c}{MCCL} & \multicolumn{1}{c}{SCED} & UniTR &
			\multicolumn{1}{c}{CONDA} &  \multicolumn{1}{c}{OURS$_{I}$} & \multicolumn{1}{c}{OURS$_{II}$} & \multicolumn{1}{c}{\multirow{3}[2]{*}{\textit{VS.}}} \\
			\multicolumn{1}{c|}{} &       & \multicolumn{1}{c|}{CVPR$_{23}$} &   \multicolumn{1}{c}{ICCV$_{21}$} & \multicolumn{1}{c}{CVPR$_{21}$} &  \multicolumn{1}{c}{CVPR$_{22}$} & \multicolumn{1}{c}{CVPR$_{23}$} & \multicolumn{1}{c}{TPAMI$_{23}$} & TPAMI$_{23}$ & \multicolumn{1}{c}{CVPR$_{23}$} & \multicolumn{1}{c}{TPAMI$_{23}$} & \multicolumn{1}{c}{AAAI$_{23}$} & \multicolumn{1}{c}{ACMM$_{23}$} &
			TMM$_{24}$ &
			\multicolumn{1}{c}{ECCV$_{24}$} &  \multicolumn{1}{c}{—} & \multicolumn{1}{c}{—} &  \\
			\multicolumn{1}{c|}{} &       & \multicolumn{1}{c|}{S+D} &  \multicolumn{1}{c}{C+D} & \multicolumn{1}{c}{D} &  \multicolumn{1}{c}{C} & \multicolumn{1}{c}{C+D} & \multicolumn{1}{c}{C+D} & D     & \multicolumn{1}{c}{S+D} & \multicolumn{1}{c}{S+D} & \multicolumn{1}{c}{S+D} & \multicolumn{1}{c}{S+D} &
			S+D     &
			\multicolumn{1}{c}{S+D}  & \multicolumn{1}{c}{C+D} & \multicolumn{1}{c}{S+D} &  \\
			\midrule
			\multicolumn{1}{c|}{\multirow{6}[2]{*}{CoCA}} & \multicolumn{1}{c|}{$S_m$↑} & 0.686  & 0.681 & 0.673 & 0.710 & 0.728 & 0.715 & \multicolumn{1}{c}{0.708} & 0.726 & 0.738 & 0.713 & 0.741 & 0.708 & \textcolor[rgb]{ 0,  1,  0}{0.763} & \textcolor[rgb]{ 0,  0,  1}{0.774} & \textcolor[rgb]{ 1,  0,  0}{\textbf{0.819}} & +5.6\% \\
			\multicolumn{1}{c|}{} & \multicolumn{1}{c|}{$E_m^{\max}$↑} & 0.760  & 0.744 & 0.760 & 0.783 & 0.800 & 0.769 & \multicolumn{1}{c}{0.791} & 0.808 & 0.814 & 0.796 & 0.828  & 0.789 & \textcolor[rgb]{ 0,  0,  1}{0.839} & \textcolor[rgb]{ 0,  1,  0}{0.829} & \textcolor[rgb]{ 1,  0,  0}{\textbf{0.871}} & +3.2\% \\
			\multicolumn{1}{c|}{} & \multicolumn{1}{c|}{$F_m^{\max}$↑} & 0.546   & 0.548 & 0.544 & 0.598 & 0.621 & 0.597 & \multicolumn{1}{c}{0.609} & 0.599 & 0.637 & 0.584 & 0.629 & 0.574 & \textcolor[rgb]{ 0,  0,  1}{0.685} & \textcolor[rgb]{ 0,  1,  0}{0.680} & \textcolor[rgb]{ 1,  0,  0}{\textbf{0.752}} & +6.7\% \\
			\multicolumn{1}{c|}{} & \multicolumn{1}{c|}{MAE↓} & 0.126  & 0.132 & 0.105 & 0.085 & 0.107 & 0.103 & \multicolumn{1}{c}{0.107} & 0.095 & \textcolor[rgb]{ 0,  1,  0}{0.081} & 0.097 & 0.084 & 0.089 & 0.089 & \textcolor[rgb]{ 0,  0,  1}{0.069} & \textcolor[rgb]{ 1,  0,  0}{\textbf{0.054}} & -2.7\% \\
			\multicolumn{1}{c|}{} & \multicolumn{1}{c|}{$E_m$↑} & 0.708   & 0.690 & 0.739 & 0.778 & 0.754 & 0.747 & —     & 0.767 & 0.783 & 0.764 & \textcolor[rgb]{ 0,  1,  0}{0.804} & 0.766 & 0.790 & \textcolor[rgb]{ 0,  0,  1}{0.813} & \textcolor[rgb]{ 1,  0,  0}{\textbf{0.830}} & +2.6\% \\
			\multicolumn{1}{c|}{} & \multicolumn{1}{c|}{$F_m$↑} & 0.510 & 0.503 & 0.531 & 0.593 & 0.590 & 0.584 & —     & 0.566 & 0.612 & 0.560 & 0.610 & 0.559 & \textcolor[rgb]{ 0,  1,  0}{0.640} & \textcolor[rgb]{ 0,  0,  1}{0.660} & \textcolor[rgb]{ 1,  0,  0}{\textbf{0.708}} & +6.8\% \\
			\midrule
			\multicolumn{1}{c|}{\multirow{6}[2]{*}{CoSOD3k}} & \multicolumn{1}{c|}{$S_m$↑} & 0.839  & 0.801 & 0.802 & 0.810 & 0.852 & 0.838 & \multicolumn{1}{c}{0.823} & 0.853 & 0.843 & 0.854 & \textcolor[rgb]{ 0,  1,  0}{0.865} & 0.852 & 0.862 & \textcolor[rgb]{ 0,  0,  1}{0.874} & \textcolor[rgb]{ 1,  0,  0}{\textbf{0.895}} & +3.0\% \\
			\multicolumn{1}{c|}{} & \multicolumn{1}{c|}{$E_m^{\max}$↑} & 0.854  & 0.840 & 0.860 & 0.874 & 0.895 & 0.890 & \multicolumn{1}{c}{0.876} & 0.911 & 0.901 & 0.902 &  \textcolor[rgb]{ 0,  0,  1}{0.923} & 0.903 & 0.911 & \textcolor[rgb]{ 0,  1,  0}{0.918} & \textcolor[rgb]{ 1,  0,  0}{\textbf{0.938}} & +1.5\% \\
			\multicolumn{1}{c|}{} & \multicolumn{1}{c|}{$F_m^{\max}$↑} & 0.813  & 0.759 & 0.778 & 0.805 & 0.836 & 0.827 & \multicolumn{1}{c}{0.808} & 0.829 & 0.834 & 0.832 & \textcolor[rgb]{ 0,  1,  0}{0.859} & 0.834 & 0.853 & \textcolor[rgb]{ 0,  0,  1}{0.868} & \textcolor[rgb]{ 1,  0,  0}{\textbf{0.893}} & +3.4\% \\
			\multicolumn{1}{c|}{} & \multicolumn{1}{c|}{MAE↓} & 0.076  & 0.096 & 0.071 & 0.067 & 0.064 & 0.060 & \multicolumn{1}{c}{0.070} & 0.062 & 0.062 & 0.062 & \textcolor[rgb]{ 0,  1,  0}{0.053} & 0.058 & 0.056 & \textcolor[rgb]{ 0,  0,  1}{0.049} & \textcolor[rgb]{ 1,  0,  0}{\textbf{0.043}} & -1.0\% \\
			\multicolumn{1}{c|}{} & \multicolumn{1}{c|}{$E_m$↑} & 0.852  & 0.824 & 0.857 & 0.871 & 0.882 & 0.886 & —     & 0.885 & 0.872 & 0.879 & \textcolor[rgb]{ 0,  1,  0}{0.900} & 0.890 & 0.893 & \textcolor[rgb]{ 0,  0,  1}{0.901} & \textcolor[rgb]{ 1,  0,  0}{\textbf{0.909}} & +0.9\% \\
			\multicolumn{1}{c|}{} & \multicolumn{1}{c|}{$F_m$↑} & 0.774  & 0.743 & 0.770 & 0.800 & 0.815 & 0.815 & —     & 0.803 & 0.813 & 0.809 & \textcolor[rgb]{ 0,  1,  0}{0.842} & 0.821 & 0.827 & \textcolor[rgb]{ 0,  0,  1}{0.850} & \textcolor[rgb]{ 1,  0,  0}{\textbf{0.860}} & +1.8\% \\
			\midrule
			\multicolumn{1}{c|}{\multirow{6}[1]{*}{CoSal2015}} & \multicolumn{1}{c|}{$S_m$↑} & 0.876  & 0.866 & 0.845 & 0.838 & 0.896 & 0.867 & \multicolumn{1}{c}{0.875} & 0.885 & 0.881 & 0.887 & 0.894 & 0.896 & \textcolor[rgb]{ 0,  1,  0}{0.900} & \textcolor[rgb]{ 0,  0,  1}{0.911} & \textcolor[rgb]{ 1,  0,  0}{\textbf{0.927}} & +2.7\% \\
			\multicolumn{1}{c|}{} & \multicolumn{1}{c|}{$E_m^{\max}$↑} & 0.914  & 0.906 & 0.887 & 0.892 & 0.933 & 0.912 & \multicolumn{1}{c}{0.922} & 0.933 & 0.924 & 0.923 & 0.938 &  \textcolor[rgb]{ 0,  1,  0}{0.940} &  \textcolor[rgb]{ 0,  0,  1}{0.944} & \textcolor[rgb]{ 0,  0,  1}{0.944} & \textcolor[rgb]{ 1,  0,  0}{\textbf{0.962}} & +1.8\% \\
			\multicolumn{1}{c|}{} & \multicolumn{1}{c|}{$F_m^{\max}$↑} & 0.874  & 0.862 & 0.847 & 0.856 & 0.903 & 0.882 & \multicolumn{1}{c}{0.889} & 0.882 & 0.891 & 0.887 & \textcolor[rgb]{ 0,  1,  0}{0.908} & 0.902 & \textcolor[rgb]{ 0,  1,  0}{0.908} & \textcolor[rgb]{ 0,  0,  1}{0.920} & \textcolor[rgb]{ 1,  0,  0}{\textbf{0.941}} & +3.3\% \\
			\multicolumn{1}{c|}{} & \multicolumn{1}{c|}{MAE↓} & 0.068  & 0.064 & 0.068 & 0.067 & 0.047 & 0.055 & \multicolumn{1}{c}{0.047} & 0.053 & 0.056 & 0.054 & 0.045 & \textcolor[rgb]{ 0,  1,  0}{0.041} & 0.045 & \textcolor[rgb]{ 0,  0,  1}{0.037} & \textcolor[rgb]{ 1,  0,  0}{\textbf{0.030}} & -1.1\% \\
			\multicolumn{1}{c|}{} & \multicolumn{1}{c|}{$E_m$↑} & 0.881  & 0.874 & 0.884 & 0.888 & 0.922 & 0.907 & —     & 0.913 & 0.902 & 0.907 & 0.924 & \textcolor[rgb]{ 0,  1,  0}{0.928} & 0.923 & \textcolor[rgb]{ 0,  0,  1}{0.933} & \textcolor[rgb]{ 1,  0,  0}{\textbf{0.944}} & +1.6\% \\
			\multicolumn{1}{c|}{} & \multicolumn{1}{c|}{$F_m$↑} & 0.836  & 0.826 & 0.838 & 0.850 & 0.880 & 0.869 & —     & 0.856 & 0.870 & 0.868 & \textcolor[rgb]{ 0,  1,  0}{0.892} & 0.887 & 0.887 & \textcolor[rgb]{ 0,  0,  1}{0.902} & \textcolor[rgb]{ 1,  0,  0}{\textbf{0.915}} & +2.3\% \\
			\midrule
			\multicolumn{2}{c|}{Tunable Param. (M)} & 3.7   & 392.85 & 280.36   & 142.3 & 40.4  & 20    & \multicolumn{1}{c}{393.21} & 52.3  & 18.4  & 27.1  & 156.7 & 146.6 & 24.1 & 4.94   & 4.94   &  \\
			\midrule
			\multicolumn{2}{c|}{Model Size (MB)} & 14.1   & 1498.7 & 541.7  & 542.9 & 154.4 & 228.3 & —     & 199.7 & 70.4  & 104.5 & 1750  & 541 & 139.5 & 19    & 19    &  \\
			\bottomrule
		\end{tabular}%
	}
	\vspace{-0.05in}
\end{table*}%
\begin{table*}[!t]
	\centering
	\normalsize
	\tabcolsep=0.2cm
	\renewcommand{\arraystretch}{0.95}
	\caption{Ablation analysis on the main components of our VCP architecture. “BL” represents the baseline, which utilizes solely the frozen foundation model and the tunable FPN-like decoder. $P_{Hand}$ represents $P_{Em}$+$P_{Hand}$ and $P_{Co}$ represents $P_{Em}$+$P_{Co}$.}  \label{tab-2}
	\vspace{-0.1in}
	\scalebox{0.85}{
		\begin{tabular}{cccc|c|c|ccc|ccc|ccc}
			\toprule
			\multicolumn{4}{c|}{Combination} & \multicolumn{1}{c|}{\multirow{2}[2]{*}{\makecell[c]{Tunable \\ Param. (M)}}} &
			\multicolumn{1}{c|}{\multirow{2}[2]{*}{\makecell[c]{Model Size \\ (MB)}}} & \multicolumn{3}{c|}{COCA} & \multicolumn{3}{c|}{CoSOD3k} & \multicolumn{3}{c}{CoSal2015} \\
			\cmidrule{1-4}\cmidrule{7-15}    BL    & PH    & $P_{Hand}$ & $P_{Co}$  &     &     & $S_m$↑   & $F_m^{\max}$↑ & MAE↓  & $S_m$↑   & $F_m^{\max}$↑ & MAE↓  & $S_m$↑   & $F_m^{\max}$↑ & MAE↓ \\
			\midrule
			\checkmark     &       &       &       & 0.14 & 0.52 & 0.635 & 0.470 & 0.138 & 0.777 & 0.749 & 0.099 & 0.813 & 0.814 & 0.099 \\
			\checkmark     & \checkmark     &       &       & 1.49   & 5.73  & 0.659 & 0.498 & 0.125 & 0.832 & 0.795 & 0.073 & 0.866 & 0.862 & 0.063 \\
			\checkmark     & \checkmark     & \checkmark     &       & 2.04  & 7.87  & 0.704 & 0.561 & 0.111 & 0.852 & 0.831 & 0.067 & 0.894 & 0.893 & 0.054 \\
			\checkmark     & \checkmark     &       & \checkmark     & 4.59  & 17.67  & 0.763 & 0.662 & 0.073 & 0.873 & \textcolor[rgb]{ 1,  0,  0}{\textbf{0.870}} & 0.052 & \textcolor[rgb]{ 1,  0,  0}{\textbf{0.911}} & \textcolor[rgb]{ 1,  0,  0}{\textbf{0.923}} & 0.043 \\
			\checkmark     & \checkmark     & \multicolumn{2}{c|}{Addition} & 4.72 & 18.14  & 0.755 & 0.646 & 0.080 & 0.873 & 0.861 & 0.048 & 0.910 & 0.916 & \textcolor[rgb]{ 1,  0,  0}{\textbf{0.036}} \\
			\midrule
			\checkmark     & \checkmark     & \multicolumn{2}{c|}{Concatenation} & 4.94  & 19.02  & \textcolor[rgb]{ 1,  0,  0}{\textbf{0.774}} & \textcolor[rgb]{ 1,  0,  0}{\textbf{0.680}} & \textcolor[rgb]{ 1,  0,  0}{\textbf{0.069}} & \textcolor[rgb]{ 1,  0,  0}{\textbf{0.874}} & 0.868 & \textcolor[rgb]{ 1,  0,  0}{\textbf{0.049}} & \textcolor[rgb]{ 1,  0,  0}{\textbf{0.911}} & 0.920 & 0.037 \\
			\bottomrule
		\end{tabular}%
	}
	\vspace{-0.15in}
\end{table*}%

\vspace{-0.15in}
\subsection{Consensus Prompt Disperser}
To effectively leverage the obtained consensus prompts to induce the foundation model to perform well on CoSOD tasks, a concise CPD is further proposed. Firstly, the obtained consensus prompts ${P_{Co}}$ are used to further guide the consensus expression within the embedding prompts ${P_{Em}}$ through simple integration, denoted as Embedding Consensus Prompts $P_{Em}^{Co} \in {\mathbb{R}^{N \times {L_s} \times {C_r}}}$. The $P_{Em}^{Co}$ are more effective for tuning the frozen foundation model's embedding components. Additionally, inspired by some methods that incorporate handcrafted features into models and gain benefits. For example, the EVP \cite{liu2023explicit} utilizes handcrafted features derived from fast Fourier transform as Explicit Visual Prompts to achieve effective foreground segmentation. Therefore, we also consider incorporating handcrafted features ${P_{Hand}}$ as part of the visual prompts to focus on high-frequency components within individual images. However, since the handcrafted prompts used in EVP only focus on individual independent images, their effectiveness is significantly discounted when dealing with CSOD tasks that require emphasis on intra-group relevance (as demonstrated in the experimental section). Taking these considerations into account, we once again utilize the obtained consensus prompts ${P_{Co}}$ to highlight the consensus high-frequency components in the handcrafted prompts ${P_{Hand}} \in {\mathbb{R}^{N \times {L_s} \times {C_r}}}$, denoted as Handcrafted Consensus Prompts $P_{Hand}^{Co} \in {\mathbb{R}^{N \times {L_s} \times {C_r}}}$. The handcrafted prompts ${P_{Hand}}$ introduced in our method are obtained directly from the input image through predefined multi-stage overlapping patch embeddings after fast Fourier transformation.

The $P_{Em}^{Co}$ and $P_{Hand}^{Co}$ are further integrated to obtain Visual Consensus Prompts $P_{Visual}^{Co} \in {\mathbb{R}^{N \times {L_s} \times {2C_r}}}$.
\begin{equation}
\begin{array}{c}
P_{Visual}^{Co} = [{P_{Em}} + {P_{Co}},{P_{Hand}} + {P_{Co}}],
\end{array}
\end{equation}
Utilizing these visual consensus prompts to perform adaptive tunes on different depths of vision transformer layers is also crucial for inducing the foundation model to address CoSOD tasks. Therefore, multiple unshared linear layers are used to achieve adaptive tunes of different transformer layers. Then, an intra-stage shared linear layer is utilized to maintain consistency in dimensions between the prompts $P = \{ P_S^n \in {\mathbb{R}^{N \times {L_s} \times {C_s}}}\} _{s = 1}^4$ and the transformer features.
\begin{equation}
\begin{array}{c}
P_S^n = ML{P_{up}}(gelu(MLP_S^n(P_{Visual}^{Co}))).
\end{array}
\end{equation}
Finally, the obtained task-specific prompts $P$ will induce the foundation model to achieve effective and efficient CoSOD performance.

\subsection{Prediction Head and Objective Function}
The decoder of Segformer \cite{xie2021segformer} utilizes multiple linear layers to up-project multi-stage features to a unified dimension (typically higher, d=768). The multi-scale features are concatenated in the high dimension and down-projected through linear layers to generate predictions. The decoder of Segformer possesses a significant number of parameters (3.15M), which greatly increases the number of tunable parameters, contradicting the motivation to minimize tunable parameters. Therefore, a concise prediction head is designed, primarily comprising an ASPP, FPN-like, and an additional linear classifier. This prediction head accepts a smaller unified projection dimension and has fewer tunable parameters (1.49M).

The prediction head generates a final prediction map $M$ and classification predictions ${V_{cla}}$. We adopt a weighted combination ${L_{w}}$ of commonly used binary cross-entropy loss (BCE) and IOU loss to constrain the prediction map. The predicted classes are supervised using cross-entropy loss ${L_{ce}}$. Simultaneously, we utilize the saliency estimation maps generated by the CPG in four stages as initial co-salient object predictions $\{ {M^s}\} _{s = 1}^4$ to constrain them towards the final prediction targets. The overall loss can be expressed as:
\vspace{-0.19in}
\begin{equation}
\begin{array}{c}
L = \alpha {L_w}(M,GT) + \beta \sum\limits_{s = 1}^4 {{L_w}({M^s},GT)} \\ + \lambda {L_{ce}}({V_{cla}},{V_{lab}}).
\end{array}
\end{equation}
Where $\alpha$, $\beta$, and $\lambda$ are set to 10, 2, and 0.1, respectively.

\section{Experiments}
\subsection{Datasets and Evaluation Metrics}

\begin{figure*}[!t]
	\centering
	\includegraphics[width=6.7in]{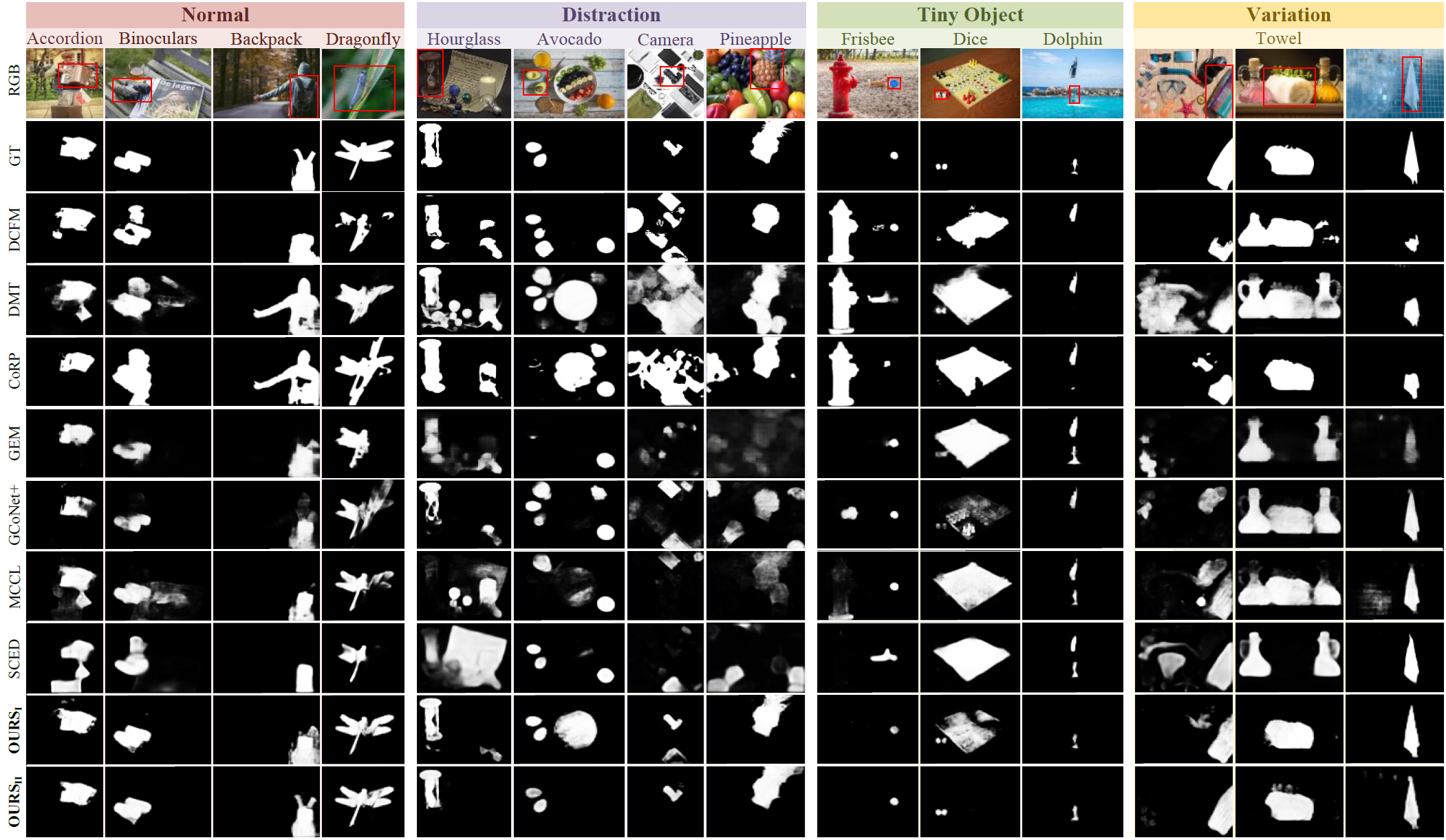}
	\vspace{-0.2in}
	\DeclareGraphicsExtensions.
	\begin{center}
		\caption{Visual comparison between our VCP and the most representative seven methods across four scenarios.}\label{fig-6}
	\end{center}
	\vspace{-0.3in}
\end{figure*}  
\begin{table*}[!t]
	\centering
	\normalsize
	\tabcolsep=0.2cm
	\renewcommand{\arraystretch}{0.95}
	\caption{Ablation on the CPD details and some auxiliary components in the architecture.}  \label{tab-3}
	\vspace{-0.1in}
	\scalebox{0.85}{
		\begin{tabular}{c|c|c|ccc|ccc|ccc}
			\toprule
			\multirow{2}[4]{*}{Settings} & \multicolumn{1}{c|}{\multirow{2}[2]{*}{\makecell[c]{Tunable \\ Param. (M)}}} &
			\multicolumn{1}{c|}{\multirow{2}[2]{*}{\makecell[c]{Model Size \\ (MB)}}} & \multicolumn{3}{c|}{COCA} & \multicolumn{3}{c|}{CoSOD3k} & \multicolumn{3}{c}{CoSal2015} \\
			\cmidrule{4-12}          &       &     & $S_m$↑   & $F_m^{\max}$↑ & MAE↓  & $S_m$↑   & $F_m^{\max}$↑ & MAE↓  & $S_m$↑   & $F_m^{\max}$↑ & MAE↓ \\
			\midrule
			w/o ${L_{ce}}({V_{cla}},{V_{lab}})$ & 4.92 & 18.93  & 0.761 & 0.653 & 0.078 & 0.872 & 0.864 & 0.051 & \textcolor[rgb]{ 1,  0,  0}{\textbf{0.911}} & 0.919 & 0.038 \\
			w/o ${{L_w}({M^s},GT)}$ & 4.94 & 19.02  & 0.742 & 0.634 & 0.096 & 0.870 & 0.853 & 0.055 & \textcolor[rgb]{ 1,  0,  0}{\textbf{0.911}} & 0.915 & \textcolor[rgb]{ 1,  0,  0}{\textbf{0.037}} \\
			w/ Segformer Head & 3.84  & 14.73  & 0.747 & 0.656 & 0.101 & 0.872 & 0.861 & 0.054 & 0.909 & 0.916 & 0.038 \\
			\midrule
			w/ Share MLP & 4.53  & 17.41  & 0.768 & 0.669 & \textcolor[rgb]{ 1,  0,  0}{\textbf{0.067}} & 0.871 & 0.865 & 0.051 & 0.905 & 0.914 & 0.042 \\
			w/ Unshare MLP & 5.78  & 22.24  & 0.763 & 0.655 & 0.076 & 0.873 & 0.865 & 0.051 & \textcolor[rgb]{ 1,  0,  0}{\textbf{0.911}} & 0.918 & 0.038 \\
			\midrule
			\textbf{OURS(Adaptive MLP)}  & 4.94 & 19.02  & \textcolor[rgb]{ 1,  0,  0}{\textbf{0.774}} & \textcolor[rgb]{ 1,  0,  0}{\textbf{0.680}} & 0.069 & \textcolor[rgb]{ 1,  0,  0}{\textbf{0.874}} & \textcolor[rgb]{ 1,  0,  0}{\textbf{0.868}} & \textcolor[rgb]{ 1,  0,  0}{\textbf{0.049}} & \textcolor[rgb]{ 1,  0,  0}{\textbf{0.911}} & \textcolor[rgb]{ 1,  0,  0}{\textbf{0.920}} & \textcolor[rgb]{ 1,  0,  0}{\textbf{0.037}} \\
			\bottomrule
		\end{tabular}%
	}
	\vspace{-0.15in}
\end{table*}%

\textbf{Datasets.} All experiments are evaluated based on the most commonly used three datasets for CoSOD tasks: CoCA \cite{zhang2020gradient}, CoSOD3k \cite{fan2021re}, and CoSal2015 \cite{zhang2015co}. The three primary training datasets widely used in CoSOD tasks are the DUT-class \cite{zhang2020gradient}, COCO-9k \cite{lin2014microsoft}, and COCO-SEG \cite{wang2019robust} datasets, and we denote them as D, C, and S, respectively. The combinations of training datasets primarily used by existing methods include D+C \cite{jin2020icnet, li2023discriminative, zhang2021summarize, zhu2023co}, as well as D+S \cite{li2024conda, wu2023co, xu2023co, zheng2023memory}. In the comparative experiments, we conduct experiments using both combinations of training data to ensure fairness of comparison. In all ablation experiments, we use the DUT-class and COCO-9k (D+C) datasets as the training set. 

\textbf{Evaluation Metrics.} Six evaluation metrics are used for comparison, including \textsl{MAE} \cite{cheng2013efficient}, Mean-Em ($E_m$), $E_m^{\max}$ \cite{Fan2018EnhancedalignmentMF}, $S_m$ \cite{fan2017structure}, Mean-Fm ($F_m$) and $F_m^{\max}$ \cite{achanta2009frequency}, which are to assess the average pixel-wise absolute difference, local and global similarity, structural similarity between the predictions and the ground truths, and the weighted harmonic mean of precision and recall, respectively. 

\subsection{Implementation Details}
The proposed method is based on the Pytorch framework while using a pre-trained model of Segformer \cite{xie2021segformer} on the ImageNet dataset, and all experiments are conducted using a single NVIDIA 3090 GPU. We randomly pick $N$ samples
from three different groups in each training batch.
\begin{equation}
\begin{array}{c}
N = min({N_{g}}(A),{N_{g}}(B),{N_{g}}(C),16).
\end{array}
\end{equation}
Where ${N_{g}}$ means the number of images in the corresponding group. For inference, all samples in each group are input at one time. We train our network using the AdamW optimizer, the initial learning rate is initially set to 5e-4, and cosine decay is applied to the learning rate. When training with D+C, we train the model for 100 epochs with a learning rate decay to 1e-4, and the total training time is around 9 hours and the inference time is around 65.3 FPS. When training with D+S, we train for 200 epochs with a learning rate decay to 1e-5. The inputs are resized into 288 × 288 for both training and inference.

\begin{table*}[t!]
	\centering
	\normalsize
	\tabcolsep=0.24cm
	\renewcommand{\arraystretch}{0.98}
	\caption{Ablation on the proposed visual consensus prompts $P_{Visual}^{Co}$. We introduce other visual prompts proposed in some parameter-efficient prompt tuning methods to replace our VCP in performing CoSOD tasks ($prompt=Adaptformer/VPT/EVP$).}  \label{tab-4}
	\vspace{-0.1in}
	\scalebox{0.82}{
		\begin{tabular}{c|c|c|ccc|ccc|ccc}
			\toprule
			\multirow{2}[4]{*}{Prompt} & \multicolumn{1}{c|}{\multirow{2}[2]{*}{\makecell[c]{Tunable \\ Param. (M)}}} &
			\multicolumn{1}{c|}{\multirow{2}[2]{*}{\makecell[c]{Model Size \\ (MB)}}} &
			\multicolumn{3}{c|}{COCA} & \multicolumn{3}{c|}{CoSOD3k} & \multicolumn{3}{c}{CoSal2015} \\
			\cmidrule{4-12}          &       &       & $S_m$↑   & $F_m^{\max}$↑ & MAE↓  & $S_m$↑   & $F_m^{\max}$↑ & MAE↓  & $S_m$↑   & $F_m^{\max}$↑ & MAE↓ \\
			\midrule
			OURS(Full-tuning) & 65.79 & 251.43 & 0.740 & 0.624 & 0.093 & 0.866 & 0.854 & 0.055 & 0.905 & 0.911 & 0.040 \\
			OURS(Only PH) & 1.49  & 5.73  & 0.659 & 0.498 & 0.125 & 0.832 & 0.795 & 0.073 & 0.866 & 0.862 & 0.063 \\
			Prompt=Adaptformer \cite{chen2022adaptformer} & 1.54  & 5.99  & 0.695 & 0.547 & 0.111 & 0.852 & 0.831 & 0.066 & 0.883 & 0.881 & 0.061 \\
			Prompt=VPT-Deep \cite{jia2022visual} & 1.60  & 6.20  & 0.700 & 0.554 & 0.106 & 0.853 & 0.832 & 0.067 & 0.889 & 0.889 & 0.060 \\
			Prompt=EVP \cite{liu2023explicit}  & 2.04  & 7.87  & 0.704 & 0.561 & 0.111 & 0.852 & 0.831 & 0.067 & 0.894 & 0.893 & 0.054 \\
			\midrule
			EVP(S+D) \cite{liu2023explicit}   & 3.70  & 14.10  & 0.686 & 0.546 & 0.126 & 0.839 & 0.813 & 0.076 & 0.876 & 0.874 & 0.068 \\
			\textbf{VCP($r=8$)}   & 3.34  & 12.89  & 0.769 & 0.665 & \textcolor[rgb]{ 1,  0,  0}{\textbf{0.069}} & 0.872 & 0.864 & 0.051 & \textcolor[rgb]{ 1,  0,  0}{\textbf{0.913}} & 0.918 & \textcolor[rgb]{ 1,  0,  0}{\textbf{0.035}} \\
			\textbf{VCP($r=4$)}   & 4.94  & 19.02  & \textcolor[rgb]{ 1,  0,  0}{\textbf{0.774}} & \textcolor[rgb]{ 1,  0,  0}{\textbf{0.680}} & \textcolor[rgb]{ 1,  0,  0}{\textbf{0.069}} & \textcolor[rgb]{ 1,  0,  0}{\textbf{0.874}} & \textcolor[rgb]{ 1,  0,  0}{\textbf{0.868}} & \textcolor[rgb]{ 1,  0,  0}{\textbf{0.049}} & 0.911 & \textcolor[rgb]{ 1,  0,  0}{\textbf{0.920}} & 0.037 \\
			\bottomrule
		\end{tabular}%
	}
	\vspace{-0.18in}
\end{table*}%
\vspace{-0.05in}

\subsection{Comparisons with State-of-the-art Methods}
To demonstrate the effectiveness of our VCP, we compare it with 13 SOTA CoSOD methods from the past three years, including SCED \cite{xu2023co}, MCCL \cite{zheng2023memory}, GCONet+ \cite{zheng2023gconet+}, GEM \cite{wu2023co}, CADC++ \cite{zhang2023cadc++}, CoPR \cite{zhu2023co}, DMT \cite{li2023discriminative}, DCFM \cite{zhang2019co}, GCoNet \cite{fan2021group}, CADC \cite{zhang2021summarize}, UniTR \cite{guo2024unitr}, CONDA \cite{li2024conda} and EVP \cite{liu2023explicit}. All methods except EVP are based on Full fine-tuning for CoSOD task. EVP addresses foreground segmentation by introducing simple visual prompts. Since there are currently no methods that utilize prompt learning for CoSOD tasks, we retrain EVP \cite{liu2023explicit} to attempt solving the CoSOD problem. This serves as compelling evidence to validate the effectiveness of our visual consensus prompts for CoSOD tasks. \textbf{Quantitative Results:} Table \ref{tab-1} presents a quantitative comparison of our VCP with the most representative works in the past three years across multiple metrics. Compared to 12 full fine-tuning CoSOD methods, our VCP demonstrates a significant performance advantage on three commonly used benchmark test sets.  Compared to the prompt-based tuning method EVP \cite{liu2023explicit}, our VCP exhibits overwhelmingly superior performance. \textbf{Qualitative Results:} Fig. \ref{fig-6} illustrates the visual comparisons between our VCP and seven representative works across four selected scenarios (“Normal”, “Distraction”, “Tiny Object” and “Variation”). Through the visual comparisons in Fig. \ref{fig-6}, it can be observed that our VCP achieves more effective localization and segmentation performance even when facing challenging cases such as multiple salient object interferences, small or diverse co-salient objects.

\subsection{Ablation Study}
We conduct ablation experiments to demonstrate the effectiveness of each component of the proposed VCP. 
All ablation experiments are conducted using the C+D dataset as the training set, tested on three mainstream benchmark datasets. 

\textbf{Effectiveness of the architecture designs.} Table \ref{tab-2} shows the performance contributions of the main components in our architecture. “BL” represents the baseline, which utilizes solely the frozen foundation model and the tunable FPN-like decoder. $P_{Hand}$ represents $P_{Em}$+$P_{Hand}$ and $P_{Co}$ represents $P_{Em}$+$P_{Co}$. “Addition” represents the addition to obtain visual consensus prompts, i.e., $P_{Visual}^{Co}=P_{Em}^{Co}+P_{Hand}^{Co}$. It can be observed that the introduced handcrafted prompts show certain improvement but the enhancement is limited, while our proposed consensus prompts significantly enhance the effectiveness of the model. 


\textbf{Effectiveness of CPD and auxiliary components.} Table \ref{tab-3} presents the implementation details of CPD and some ablations of auxiliary components in the architecture. “${L_{ce}}({V_{cla}},{V_{lab}})$” denotes the removal of the classifier loss, and “${{L_w}({M^s},GT)}$” denotes the removal of supervision for the multiscale predictions generated in CPG. Furthermore, we further execute two schemes to validate the effectiveness of CPD in performing adaptive guidance after obtaining visual consensus prompts. For different depths ($n$) of Transformer layers within a specific stage, our CPD employs an adaptive tuning scheme, i.e., shared $ML{P_{up}}$ and unshared $MLP_S^n$, while “Share MLP” and “Unshare MLP” respectively indicate whether $ML{P_{up}}$ and $MLP_S^n$ are shared or not. We employ an adaptive approach to disperse visual consensus prompts $P_{Visual}^{Co}$, achieving a balance between parameter efficiency and effective performance.

\textbf{Effectiveness of the proposed CPG.} To validate the effectiveness of the consensus prompts generated by our CPG, we replace our CPG with other visual prompts (\textsl{Adaptformer/VPT/EVP}) from state-of-the-art prompt tuning methods. As seen in Table \ref{tab-4}, our VCP significantly outperforms existing visual prompt schemes by a large margin in addressing specific CoSOD tasks, and it can comprehensively outperform our full fine-tuning scheme.

Additional experimental details and scalability validations are provided in the \textbf{supplementary materials}.

\section{Conclusion}

In this work, we propose VCP, a interaction-effective and parameter-efficient visual prompt tuning framework for CoSOD, serving as a potent alternative to existing methods based on the common architecture and the full fine-tuning paradigm. Additionally, our formulated visual consensus prompts (with minimized tunable parameters) effectively and efficiently induce the frozen pre-trained foundation model to play a crucial role in CoSOD tasks. Our CPG effectively enforces limited tunable parameters to focus on intra-group co-salient representations and generate key consensus prompts. Similarly, our CPD accomplishes its mission of adaptively inducing the foundation model using consensus prompts. Extensive experiments demonstrate that our VCP outperforms existing state-of-the-art visual prompt schemes by a large margin when addressing specific CoSOD tasks. Compared to state-of-the-art full fine-tuning methods tailored for CoSOD tasks, our VCP still maintains a comprehensive competitive advantage.
  
\noindent\textbf{Acknowledgment.}
This work is supported by the National Nature Science Foundation of China (Nos.62376186).

{
    \small
    \bibliographystyle{ieeenat_fullname}
    \bibliography{main}

\begin{thebibliography}{46}
\providecommand{\natexlab}[1]{#1}
\providecommand{\url}[1]{\texttt{#1}}
\expandafter\ifx\csname urlstyle\endcsname\relax
  \providecommand{\doi}[1]{doi: #1}\else
  \providecommand{\doi}{doi: \begingroup \urlstyle{rm}\Url}\fi

\bibitem[Achanta et~al.(2009)Achanta, Hemami, Estrada, and
  Susstrunk]{achanta2009frequency}
Radhakrishna Achanta, Sheila Hemami, Francisco Estrada, and Sabine Susstrunk.
\newblock Frequency-tuned salient region detection.
\newblock In \emph{Proc. IEEE/CVF Conf. Comput. Vis. Pattern Recognit.}, pages
  1597--1604, 2009.

\bibitem[Brown et~al.(2020)Brown, Mann, Ryder, Subbiah, Kaplan, Dhariwal,
  Neelakantan, Shyam, Sastry, Askell, et~al.]{brown2020language}
Tom Brown, Benjamin Mann, Nick Ryder, Melanie Subbiah, Jared~D Kaplan, Prafulla
  Dhariwal, Arvind Neelakantan, Pranav Shyam, Girish Sastry, Amanda Askell,
  et~al.
\newblock Language models are few-shot learners.
\newblock In \emph{Proc. Adv. Neural Inform. Process. Syst.}, pages 1877--1901,
  2020.

\bibitem[Chen et~al.(2022)Chen, Ge, Tong, Wang, Song, Wang, and
  Luo]{chen2022adaptformer}
Shoufa Chen, Chongjian Ge, Zhan Tong, Jiangliu Wang, Yibing Song, Jue Wang, and
  Ping Luo.
\newblock Adaptformer: Adapting vision transformers for scalable visual
  recognition.
\newblock In \emph{Proc. Adv. Neural Inform. Process. Syst.}, pages
  16664--16678, 2022.

\bibitem[Cheng et~al.(2013)Cheng, Warrell, Lin, Zheng, Vineet, and
  Crook]{cheng2013efficient}
Ming-Ming Cheng, Jonathan Warrell, Wen-Yan Lin, Shuai Zheng, Vibhav Vineet, and
  Nigel Crook.
\newblock Efficient salient region detection with soft image abstraction.
\newblock In \emph{Proc. IEEE/CVF Int. Conf. Comput. Vision}, pages 1529--1536,
  2013.

\bibitem[Fan et~al.(2017)Fan, Cheng, Liu, Li, and Borji]{fan2017structure}
Deng-Ping Fan, Ming-Ming Cheng, Yun Liu, Tao Li, and Ali Borji.
\newblock Structure-measure: A new way to evaluate foreground maps.
\newblock In \emph{Proc. IEEE/CVF Int. Conf. Comput. Vision}, pages 4548--4557,
  2017.

\bibitem[Fan et~al.(2018)Fan, Gong, Cao, Ren, Cheng, and
  Borji]{Fan2018EnhancedalignmentMF}
Deng-Ping Fan, Cheng Gong, Yang Cao, Bo Ren, Ming-Ming Cheng, and Ali Borji.
\newblock Enhanced-alignment measure for binary foreground map evaluation.
\newblock In \emph{Proc. Int. Joint Conf. Artif. Intell.}, pages 698--704,
  2018.

\bibitem[Fan et~al.(2021{\natexlab{a}})Fan, Li, Lin, Ji, Zhang, Cheng, Fu, and
  Shen]{fan2021re}
Deng-Ping Fan, Tengpeng Li, Zheng Lin, Ge-Peng Ji, Dingwen Zhang, Ming-Ming
  Cheng, Huazhu Fu, and Jianbing Shen.
\newblock Re-thinking co-salient object detection.
\newblock \emph{IEEE Trans. Pattern Anal. Mach. Intell.}, 44\penalty0
  (8):\penalty0 4339--4354, 2021{\natexlab{a}}.

\bibitem[Fan et~al.(2021{\natexlab{b}})Fan, Fan, Fu, Tang, Shao, and
  Tai]{fan2021group}
Qi Fan, Deng-Ping Fan, Huazhu Fu, Chi-Keung Tang, Ling Shao, and Yu-Wing Tai.
\newblock Group collaborative learning for co-salient object detection.
\newblock In \emph{Proc. IEEE/CVF Conf. Comput. Vis. Pattern Recognit.}, pages
  12288--12298, 2021{\natexlab{b}}.

\bibitem[Gong et~al.(2020)Gong, Wang, Mu, and Tian]{gong2020learning}
Guoqiang Gong, Xinghan Wang, Yadong Mu, and Qi Tian.
\newblock Learning temporal co-attention models for unsupervised video action
  localization.
\newblock In \emph{Proc. IEEE/CVF Conf. Comput. Vis. Pattern Recognit.}, pages
  9819--9828, 2020.

\bibitem[Guo et~al.(2024)Guo, Ying, Qi, and Qu]{guo2024unitr}
Ruohao Guo, Xianghua Ying, Yanyu Qi, and Liao Qu.
\newblock Unitr: A unified transformer-based framework for co-object and
  multi-modal saliency detection.
\newblock \emph{IEEE Trans. Multimedia}, 26:\penalty0 7622--7635, 2024.

\bibitem[Huang et~al.(2008)Huang, Guo, and Zhang]{huang2008detection}
Hailing Huang, Weiqiang Guo, and Yu Zhang.
\newblock Detection of copy-move forgery in digital images using sift
  algorithm.
\newblock In \emph{Proc. IEEE Pacific-Asia Workshop Comput. Intell. Ind.
  Appl.}, pages 272--276, 2008.

\bibitem[Jia et~al.(2022)Jia, Tang, Chen, Cardie, Belongie, Hariharan, and
  Lim]{jia2022visual}
Menglin Jia, Luming Tang, Bor-Chun Chen, Claire Cardie, Serge Belongie, Bharath
  Hariharan, and Ser-Nam Lim.
\newblock Visual prompt tuning.
\newblock In \emph{Proc. Eur. Conf. Comput. Vision}, pages 709--727. Springer,
  2022.

\bibitem[Jiang et~al.(2020)Jiang, Xu, Araki, and Neubig]{jiang2020can}
Zhengbao Jiang, Frank~F Xu, Jun Araki, and Graham Neubig.
\newblock How can we know what language models know?
\newblock \emph{Trans. Assoc. Comput. Linguistics}, 8:\penalty0 423--438, 2020.

\bibitem[Jin et~al.(2020)Jin, Xu, Cheng, Zhang, and Guo]{jin2020icnet}
Wen-Da Jin, Jun Xu, Ming-Ming Cheng, Yi Zhang, and Wei Guo.
\newblock Icnet: Intra-saliency correlation network for co-saliency detection.
\newblock In \emph{Proc. Adv. Neural Inform. Process. Syst.}, pages
  18749--18759, 2020.

\bibitem[Lester et~al.(2021)Lester, Al-Rfou, and Constant]{lester2021power}
Brian Lester, Rami Al-Rfou, and Noah Constant.
\newblock The power of scale for parameter-efficient prompt tuning.
\newblock In \emph{Proc. Conf. Empirical Methods Nat. Lang. Process.}, pages
  3045--3059, 2021.

\bibitem[Li et~al.(2023)Li, Han, Zhang, Liu, Khan, Cholakkal, Anwer, and
  Khan]{li2023discriminative}
Long Li, Junwei Han, Ni Zhang, Nian Liu, Salman Khan, Hisham Cholakkal,
  Rao~Muhammad Anwer, and Fahad~Shahbaz Khan.
\newblock Discriminative co-saliency and background mining transformer for
  co-salient object detection.
\newblock In \emph{Proc. IEEE/CVF Conf. Comput. Vis. Pattern Recognit.}, pages
  7247--7256, 2023.

\bibitem[Li et~al.(2024)Li, Liu, Zhang, Li, Khan, Anwer, Cholakkal, Han, and
  Khan]{li2024conda}
Long Li, Nian Liu, Dingwen Zhang, Zhongyu Li, Salman Khan, Rao Anwer, Hisham
  Cholakkal, Junwei Han, and Fahad~Shahbaz Khan.
\newblock Conda: Condensed deep association learning for co-salient object
  detection.
\newblock In \emph{Proc. Eur. Conf. Comput. Vision}, 2024.

\bibitem[Li and Liang(2021)]{li2021prefix}
Xiang~Lisa Li and Percy Liang.
\newblock Prefix-tuning: Optimizing continuous prompts for generation.
\newblock In \emph{Proc. Annu. Meet. Assoc. Comput. Linguistics Int. Joint
  Conf. Nat. Lang. Process.}, pages 4582--4597, 2021.

\bibitem[Lin et~al.(2014)Lin, Maire, Belongie, Hays, Perona, Ramanan,
  Doll{\'a}r, and Zitnick]{lin2014microsoft}
Tsung-Yi Lin, Michael Maire, Serge Belongie, James Hays, Pietro Perona, Deva
  Ramanan, Piotr Doll{\'a}r, and C~Lawrence Zitnick.
\newblock Microsoft coco: Common objects in context.
\newblock In \emph{Proc. Eur. Conf. Comput. Vision}, pages 740--755. Springer,
  2014.

\bibitem[Liu et~al.(2023)Liu, Shen, Pun, and Cun]{liu2023explicit}
Weihuang Liu, Xi Shen, Chi-Man Pun, and Xiaodong Cun.
\newblock Explicit visual prompting for low-level structure segmentations.
\newblock In \emph{Proc. IEEE/CVF Conf. Comput. Vis. Pattern Recognit.}, pages
  19434--19445, 2023.

\bibitem[Liu et~al.(2021)Liu, Ji, Fu, Tam, Du, Yang, and Tang]{liu2021p}
Xiao Liu, Kaixuan Ji, Yicheng Fu, Weng~Lam Tam, Zhengxiao Du, Zhilin Yang, and
  Jie Tang.
\newblock P-tuning v2: Prompt tuning can be comparable to fine-tuning
  universally across scales and tasks.
\newblock \emph{arXiv preprint arXiv:2110.07602}, 2021.

\bibitem[Luo et~al.(2024)Luo, Liu, Zhao, Yang, Zhang, Fan, Khan, and
  Han]{luo2024vscode}
Ziyang Luo, Nian Liu, Wangbo Zhao, Xuguang Yang, Dingwen Zhang, Deng-Ping Fan,
  Fahad Khan, and Junwei Han.
\newblock Vscode: General visual salient and camouflaged object detection with
  2d prompt learning.
\newblock In \emph{Proc. IEEE/CVF Conf. Comput. Vis. Pattern Recognit.}, pages
  17169--17180, 2024.

\bibitem[Nie et~al.(2023)Nie, Ni, Chang, Meng, Huo, Xiang, and
  Tian]{nie2023pro}
Xing Nie, Bolin Ni, Jianlong Chang, Gaofeng Meng, Chunlei Huo, Shiming Xiang,
  and Qi Tian.
\newblock Pro-tuning: Unified prompt tuning for vision tasks.
\newblock \emph{IEEE Trans. Circuits Syst. Video Technol.}, 2023.

\bibitem[Popescu and Farid(2005)]{popescu2005exposing}
Alin~C Popescu and Hany Farid.
\newblock Exposing digital forgeries by detecting traces of resampling.
\newblock \emph{IEEE Trans. Signal Process.}, 53\penalty0 (2):\penalty0
  758--767, 2005.

\bibitem[Radford et~al.(2018)Radford, Narasimhan, Salimans, Sutskever,
  et~al.]{radford2018improving}
Alec Radford, Karthik Narasimhan, Tim Salimans, Ilya Sutskever, et~al.
\newblock Improving language understanding by generative pre-training.
\newblock 2018.

\bibitem[Radford et~al.(2019)Radford, Wu, Child, Luan, Amodei, Sutskever,
  et~al.]{radford2019language}
Alec Radford, Jeffrey Wu, Rewon Child, David Luan, Dario Amodei, Ilya
  Sutskever, et~al.
\newblock Language models are unsupervised multitask learners.
\newblock \emph{OpenAI blog}, 1\penalty0 (8):\penalty0 9, 2019.

\bibitem[Shin et~al.(2020)Shin, Razeghi, Logan~IV, Wallace, and
  Singh]{shin2020autoprompt}
Taylor Shin, Yasaman Razeghi, Robert~L Logan~IV, Eric Wallace, and Sameer
  Singh.
\newblock Autoprompt: Eliciting knowledge from language models with
  automatically generated prompts.
\newblock In \emph{Proc. Conf. Empirical Methods Nat. Lang. Process.}, pages
  4222--4235, 2020.

\bibitem[Wang et~al.(2019)Wang, Zha, Liu, and Xie]{wang2019robust}
Chong Wang, Zheng-Jun Zha, Dong Liu, and Hongtao Xie.
\newblock Robust deep co-saliency detection with group semantic.
\newblock In \emph{Proc. AAAI Conf. Artif. Intell.}, pages 8917--8924, 2019.

\bibitem[Wang et~al.(2024)Wang, Yu, Zhang, and Han]{wang2024single}
Jie Wang, Nana Yu, Zihao Zhang, and Yahong Han.
\newblock Single-group generalized rgb and rgb-d co-salient object detection.
\newblock \emph{IEEE Trans. Circuits Syst. Video Technol.}, 2024.

\bibitem[Wang et~al.(2021)Wang, Lai, Fu, Shen, Ling, and Yang]{wang2021salient}
Wenguan Wang, Qiuxia Lai, Huazhu Fu, Jianbing Shen, Haibin Ling, and Ruigang
  Yang.
\newblock Salient object detection in the deep learning era: An in-depth
  survey.
\newblock \emph{IEEE Trans. Pattern Anal. Mach. Intell.}, 44\penalty0
  (6):\penalty0 3239--3259, 2021.

\bibitem[Wu et~al.(2023)Wu, Song, Liu, Zhang, and Liu]{wu2023co}
Yang Wu, Huihui Song, Bo Liu, Kaihua Zhang, and Dong Liu.
\newblock Co-salient object detection with uncertainty-aware group
  exchange-masking.
\newblock In \emph{Proc. IEEE/CVF Conf. Comput. Vis. Pattern Recognit.}, pages
  19639--19648, 2023.

\bibitem[Xie et~al.(2021)Xie, Wang, Yu, Anandkumar, Alvarez, and
  Luo]{xie2021segformer}
Enze Xie, Wenhai Wang, Zhiding Yu, Anima Anandkumar, Jose~M Alvarez, and Ping
  Luo.
\newblock Segformer: Simple and efficient design for semantic segmentation with
  transformers.
\newblock In \emph{Proc. Adv. Neural Inform. Process. Syst.}, pages
  12077--12090, 2021.

\bibitem[Xu and Mu(2023)]{xu2023co}
Peiran Xu and Yadong Mu.
\newblock Co-salient object detection with semantic-level consensus extraction
  and dispersion.
\newblock In \emph{Proc. ACM Int. Conf. Multimedia}, pages 2744--2755, 2023.

\bibitem[Yang et~al.(2022)Yang, Li, Ma, Yang, and Yan]{yang2022co}
Xi Yang, Shaoyi Li, Jun Ma, Jun-yan Yang, and Jie Yan.
\newblock Co-saliency-regularized correlation filter for object tracking.
\newblock \emph{Signal Processing: Image Communication}, 103:\penalty0 116655,
  2022.

\bibitem[Yu et~al.(2022)Yu, Xiao, Zhang, and Lim]{yu2022democracy}
Siyue Yu, Jimin Xiao, Bingfeng Zhang, and Eng~Gee Lim.
\newblock Democracy does matter: Comprehensive feature mining for co-salient
  object detection.
\newblock In \emph{Proc. IEEE/CVF Conf. Comput. Vis. Pattern Recognit.}, pages
  979--988, 2022.

\bibitem[Zhang et~al.(2015)Zhang, Han, Li, and Wang]{zhang2015co}
Dingwen Zhang, Junwei Han, Chao Li, and Jingdong Wang.
\newblock Co-saliency detection via looking deep and wide.
\newblock In \emph{Proc. IEEE/CVF Conf. Comput. Vis. Pattern Recognit.}, pages
  2994--3002, 2015.

\bibitem[Zhang et~al.(2016)Zhang, Han, Li, Wang, and Li]{zhang2016detection}
Dingwen Zhang, Junwei Han, Chao Li, Jingdong Wang, and Xuelong Li.
\newblock Detection of co-salient objects by looking deep and wide.
\newblock \emph{Int. J. Comput. Vis.}, 120:\penalty0 215--232, 2016.

\bibitem[Zhang et~al.(2019)Zhang, Li, Liu, and Liu]{zhang2019co}
Kaihua Zhang, Tengpeng Li, Bo Liu, and Qingshan Liu.
\newblock Co-saliency detection via mask-guided fully convolutional networks
  with multi-scale label smoothing.
\newblock In \emph{Proc. IEEE/CVF Conf. Comput. Vis. Pattern Recognit.}, pages
  3095--3104, 2019.

\bibitem[Zhang et~al.(2022)Zhang, Wu, Dong, Liu, Liu, and Liu]{zhang2022deep}
Kaihua Zhang, Yang Wu, Mingliang Dong, Bo Liu, Dong Liu, and Qingshan Liu.
\newblock Deep object co-segmentation and co-saliency detection via high-order
  spatial-semantic network modulation.
\newblock \emph{IEEE Trans. Multimedia}, 25:\penalty0 5733--5746, 2022.

\bibitem[Zhang et~al.(2021)Zhang, Han, Liu, and Shao]{zhang2021summarize}
Ni Zhang, Junwei Han, Nian Liu, and Ling Shao.
\newblock Summarize and search: Learning consensus-aware dynamic convolution
  for co-saliency detection.
\newblock In \emph{Proc. IEEE/CVF Conf. Comput. Vis. Pattern Recognit.}, pages
  4167--4176, 2021.

\bibitem[Zhang et~al.(2023)Zhang, Liu, Nan, and Han]{zhang2023cadc++}
Ni Zhang, Nian Liu, Fang Nan, and Junwei Han.
\newblock Cadc++: Advanced consensus-aware dynamic convolution for co-salient
  object detection.
\newblock \emph{IEEE Trans. Pattern Anal. Mach. Intell.}, 46\penalty0
  (5):\penalty0 2741--2757, 2023.

\bibitem[Zhang et~al.(2020)Zhang, Jin, Xu, and Cheng]{zhang2020gradient}
Zhao Zhang, Wenda Jin, Jun Xu, and Ming-Ming Cheng.
\newblock Gradient-induced co-saliency detection.
\newblock In \emph{Proc. Eur. Conf. Comput. Vision}, pages 455--472. Springer,
  2020.

\bibitem[Zheng et~al.(2023{\natexlab{a}})Zheng, Fu, Fan, Fan, Qin, Tai, Tang,
  and Van~Gool]{zheng2023gconet+}
Peng Zheng, Huazhu Fu, Deng-Ping Fan, Qi Fan, Jie Qin, Yu-Wing Tai, Chi-Keung
  Tang, and Luc Van~Gool.
\newblock Gconet+: A stronger group collaborative co-salient object detector.
\newblock \emph{IEEE Trans. Pattern Anal. Mach. Intell.}, 45\penalty0
  (9):\penalty0 10929--10946, 2023{\natexlab{a}}.

\bibitem[Zheng et~al.(2023{\natexlab{b}})Zheng, Qin, Wang, Xiang, and
  Xiong]{zheng2023memory}
Peng Zheng, Jie Qin, Shuo Wang, Tian-Zhu Xiang, and Huan Xiong.
\newblock Memory-aided contrastive consensus learning for co-salient object
  detection.
\newblock In \emph{Proc. AAAI Conf. Artif. Intell.}, pages 3687--3695,
  2023{\natexlab{b}}.

\bibitem[Zhu et~al.(2023{\natexlab{a}})Zhu, Lai, Chen, Wang, and
  Lu]{zhu2023visual}
Jiawen Zhu, Simiao Lai, Xin Chen, Dong Wang, and Huchuan Lu.
\newblock Visual prompt multi-modal tracking.
\newblock In \emph{Proc. IEEE/CVF Conf. Comput. Vis. Pattern Recognit.}, pages
  9516--9526, 2023{\natexlab{a}}.

\bibitem[Zhu et~al.(2023{\natexlab{b}})Zhu, Zhang, Lin, Sun, and
  Cheng]{zhu2023co}
Ziyue Zhu, Zhao Zhang, Zheng Lin, Xing Sun, and Ming-Ming Cheng.
\newblock Co-salient object detection with co-representation purification.
\newblock \emph{IEEE Trans. Pattern Anal. Mach. Intell.}, 45\penalty0
  (7):\penalty0 8193--8205, 2023{\natexlab{b}}.

\end{thebibliography}


\begin{thebibliography}{21}
\providecommand{\natexlab}[1]{#1}
\providecommand{\url}[1]{\texttt{#1}}
\expandafter\ifx\csname urlstyle\endcsname\relax
  \providecommand{\doi}[1]{doi: #1}\else
  \providecommand{\doi}{doi: \begingroup \urlstyle{rm}\Url}\fi

\bibitem[Cong et~al.(2018)Cong, Lei, Fu, Huang, Cao, and Ling]{cong2018hscs}
Runmin Cong, Jianjun Lei, Huazhu Fu, Qingming Huang, Xiaochun Cao, and Nam
  Ling.
\newblock Hscs: Hierarchical sparsity based co-saliency detection for rgbd
  images.
\newblock \emph{IEEE Trans. Multimedia}, 21\penalty0 (7):\penalty0 1660--1671,
  2018.

\bibitem[Fan et~al.(2021{\natexlab{a}})Fan, Li, Lin, Ji, Zhang, Cheng, Fu, and
  Shen]{fan2021re}
Deng-Ping Fan, Tengpeng Li, Zheng Lin, Ge-Peng Ji, Dingwen Zhang, Ming-Ming
  Cheng, Huazhu Fu, and Jianbing Shen.
\newblock Re-thinking co-salient object detection.
\newblock \emph{IEEE Trans. Pattern Anal. Mach. Intell.}, 44\penalty0
  (8):\penalty0 4339--4354, 2021{\natexlab{a}}.

\bibitem[Fan et~al.(2021{\natexlab{b}})Fan, Fan, Fu, Tang, Shao, and
  Tai]{fan2021group}
Qi Fan, Deng-Ping Fan, Huazhu Fu, Chi-Keung Tang, Ling Shao, and Yu-Wing Tai.
\newblock Group collaborative learning for co-salient object detection.
\newblock In \emph{Proc. IEEE/CVF Conf. Comput. Vis. Pattern Recognit.}, pages
  12288--12298, 2021{\natexlab{b}}.

\bibitem[Fu et~al.(2013)Fu, Cao, and Tu]{fu2013cluster}
Huazhu Fu, Xiaochun Cao, and Zhuowen Tu.
\newblock Cluster-based co-saliency detection.
\newblock \emph{IEEE Trans. Image Process.}, 22\penalty0 (10):\penalty0
  3766--3778, 2013.

\bibitem[Fu et~al.(2015)Fu, Xu, Lin, and Liu]{fu2015object}
Huazhu Fu, Dong Xu, Stephen Lin, and Jiang Liu.
\newblock Object-based rgbd image co-segmentation with mutex constraint.
\newblock In \emph{Proc. IEEE/CVF Conf. Comput. Vis. Pattern Recognit.}, pages
  4428--4436, 2015.

\bibitem[Jin et~al.(2020)Jin, Xu, Cheng, Zhang, and Guo]{jin2020icnet}
Wen-Da Jin, Jun Xu, Ming-Ming Cheng, Yi Zhang, and Wei Guo.
\newblock Icnet: Intra-saliency correlation network for co-saliency detection.
\newblock In \emph{Proc. Adv. Neural Inform. Process. Syst.}, pages
  18749--18759, 2020.

\bibitem[Li et~al.(2023)Li, Han, Zhang, Liu, Khan, Cholakkal, Anwer, and
  Khan]{li2023discriminative}
Long Li, Junwei Han, Ni Zhang, Nian Liu, Salman Khan, Hisham Cholakkal,
  Rao~Muhammad Anwer, and Fahad~Shahbaz Khan.
\newblock Discriminative co-saliency and background mining transformer for
  co-salient object detection.
\newblock In \emph{Proc. IEEE/CVF Conf. Comput. Vis. Pattern Recognit.}, pages
  7247--7256, 2023.

\bibitem[Lin et~al.(2014)Lin, Maire, Belongie, Hays, Perona, Ramanan,
  Doll{\'a}r, and Zitnick]{lin2014microsoft}
Tsung-Yi Lin, Michael Maire, Serge Belongie, James Hays, Pietro Perona, Deva
  Ramanan, Piotr Doll{\'a}r, and C~Lawrence Zitnick.
\newblock Microsoft coco: Common objects in context.
\newblock In \emph{Proc. Eur. Conf. Comput. Vision}, pages 740--755. Springer,
  2014.

\bibitem[Liu et~al.(2023)Liu, Shen, Pun, and Cun]{liu2023explicit}
Weihuang Liu, Xi Shen, Chi-Man Pun, and Xiaodong Cun.
\newblock Explicit visual prompting for low-level structure segmentations.
\newblock In \emph{Proc. IEEE/CVF Conf. Comput. Vis. Pattern Recognit.}, pages
  19434--19445, 2023.

\bibitem[Ranftl et~al.(2021)Ranftl, Bochkovskiy, and Koltun]{ranftl2021vision}
Ren{\'e} Ranftl, Alexey Bochkovskiy, and Vladlen Koltun.
\newblock Vision transformers for dense prediction.
\newblock In \emph{Proc. IEEE/CVF Int. Conf. Comput. Vision}, pages
  12179--12188, 2021.

\bibitem[Wang et~al.(2019)Wang, Zha, Liu, and Xie]{wang2019robust}
Chong Wang, Zheng-Jun Zha, Dong Liu, and Hongtao Xie.
\newblock Robust deep co-saliency detection with group semantic.
\newblock In \emph{Proc. AAAI Conf. Artif. Intell.}, pages 8917--8924, 2019.

\bibitem[Wang et~al.(2022)Wang, Xie, Li, Fan, Song, Liang, Lu, Luo, and
  Shao]{wang2022pvt}
Wenhai Wang, Enze Xie, Xiang Li, Deng-Ping Fan, Kaitao Song, Ding Liang, Tong
  Lu, Ping Luo, and Ling Shao.
\newblock Pvt v2: Improved baselines with pyramid vision transformer.
\newblock \emph{Comput. Vis. Media}, 8\penalty0 (3):\penalty0 415--424, 2022.

\bibitem[Wu et~al.(2023)Wu, Song, Liu, Zhang, and Liu]{wu2023co}
Yang Wu, Huihui Song, Bo Liu, Kaihua Zhang, and Dong Liu.
\newblock Co-salient object detection with uncertainty-aware group
  exchange-masking.
\newblock In \emph{Proc. IEEE/CVF Conf. Comput. Vis. Pattern Recognit.}, pages
  19639--19648, 2023.

\bibitem[Xu and Mu(2023)]{xu2023co}
Peiran Xu and Yadong Mu.
\newblock Co-salient object detection with semantic-level consensus extraction
  and dispersion.
\newblock In \emph{Proc. ACM Int. Conf. Multimedia}, pages 2744--2755, 2023.

\bibitem[Zhang et~al.(2015)Zhang, Han, Li, and Wang]{zhang2015co}
Dingwen Zhang, Junwei Han, Chao Li, and Jingdong Wang.
\newblock Co-saliency detection via looking deep and wide.
\newblock In \emph{Proc. IEEE/CVF Conf. Comput. Vis. Pattern Recognit.}, pages
  2994--3002, 2015.

\bibitem[Zhang et~al.(2016)Zhang, Han, Li, Wang, and Li]{zhang2016detection}
Dingwen Zhang, Junwei Han, Chao Li, Jingdong Wang, and Xuelong Li.
\newblock Detection of co-salient objects by looking deep and wide.
\newblock \emph{Int. J. Comput. Vis.}, 120:\penalty0 215--232, 2016.

\bibitem[Zhang et~al.(2021)Zhang, Han, Liu, and Shao]{zhang2021summarize}
Ni Zhang, Junwei Han, Nian Liu, and Ling Shao.
\newblock Summarize and search: Learning consensus-aware dynamic convolution
  for co-saliency detection.
\newblock In \emph{Proc. IEEE/CVF Conf. Comput. Vis. Pattern Recognit.}, pages
  4167--4176, 2021.

\bibitem[Zhang et~al.(2022)Zhang, Han, and Liu]{zhang2022learning}
Ni Zhang, Junwei Han, and Nian Liu.
\newblock Learning implicit class knowledge for rgb-d co-salient object
  detection with transformers.
\newblock \emph{IEEE Trans. Image Process.}, 31:\penalty0 4556--4570, 2022.

\bibitem[Zhang et~al.(2020)Zhang, Jin, Xu, and Cheng]{zhang2020gradient}
Zhao Zhang, Wenda Jin, Jun Xu, and Ming-Ming Cheng.
\newblock Gradient-induced co-saliency detection.
\newblock In \emph{Proc. Eur. Conf. Comput. Vision}, pages 455--472. Springer,
  2020.

\bibitem[Zheng et~al.(2023)Zheng, Qin, Wang, Xiang, and Xiong]{zheng2023memory}
Peng Zheng, Jie Qin, Shuo Wang, Tian-Zhu Xiang, and Huan Xiong.
\newblock Memory-aided contrastive consensus learning for co-salient object
  detection.
\newblock In \emph{Proc. AAAI Conf. Artif. Intell.}, pages 3687--3695, 2023.

\bibitem[Zhu et~al.(2023)Zhu, Zhang, Lin, Sun, and Cheng]{zhu2023co}
Ziyue Zhu, Zhao Zhang, Zheng Lin, Xing Sun, and Ming-Ming Cheng.
\newblock Co-salient object detection with co-representation purification.
\newblock \emph{IEEE Trans. Pattern Anal. Mach. Intell.}, 45\penalty0
  (7):\penalty0 8193--8205, 2023.

\end{thebibliography}
}


\end{document}


	\title{\textsl{Supplementary Materials:} \\\vspace{0.4cm} Visual Consensus Prompting for Co-Salient Object Detection}  
	
	\maketitle
	\thispagestyle{empty}
	\appendix
	
	\begin{figure*}[ht]
		\centering
		\includegraphics[width=7in]{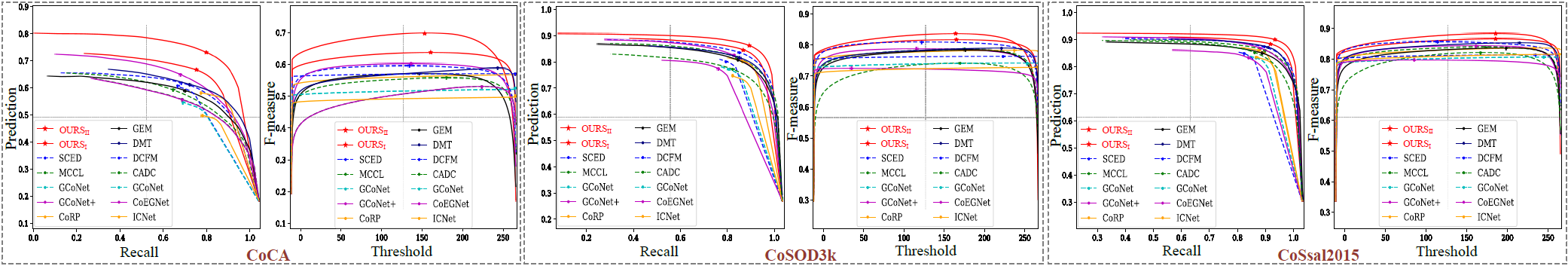}
		\vspace{-0.25in}
		\DeclareGraphicsExtensions.
		\begin{center}
			\caption*{\textbf{Fig. S1}: Precision-Recall and $F_m$-Threshold curves are compared between our VCP and 12 SOTA methods on three test sets.}\label{fig-s1}
		\end{center}
		\vspace{-0.32in}
	\end{figure*}
	
	Additional supplementary materials are provided to complement the necessary experimental details and further demonstrate the effectiveness and efficiency of our VCP. The materials mainly include: A. Additional Experimental Details. B. Ablation Analysis on Architecture Configuration and Hyperparameters. C. Some alternative schemes of VCP with more efficient parameters. D. Generalization verification on the bi-modal RGB-D Co-salient object detection (CoSOD).

	\section{Additional Experimental Details.}
	
	\textbf{Datasets.} All experiments are evaluated based on the most commonly used three datasets for CoSOD tasks: CoCA \cite{zhang2020gradient}, CoSOD3k \cite{fan2021re}, and CoSal2015 \cite{zhang2015co}. The CoCA \cite{zhang2020gradient} dataset consists of 80 categories with 1295 images. Each image in the CoCA dataset contains extraneous salient objects apart from co-salient objects, and there are many data categories in this dataset that are not encountered in the training data. Therefore, the CoCA dataset is the most challenging and best reflects the robustness and generalization of each method. Cosal2015 \cite{zhang2015co} contains 2015 images of 50 categories, and each group has one or more challenging issues such as complex environments, object appearance variations, occlusion, and background clutters. CoSOD3k \cite{fan2021re} is the largest CoSOD evaluation dataset so far, which contains 160 groups and totally 3,316 images. 
	
	The three primary training datasets widely used in CoSOD tasks are the DUT-class \cite{zhang2020gradient}, COCO-9k \cite{lin2014microsoft}, and COCO-SEG \cite{wang2019robust} datasets, and we denote them as D, C, and S, respectively. The DUTS-class \cite{zhang2020gradient} dataset consists of 291 categories with 8,250 images. The COCO-9k \cite{lin2014microsoft} dataset consists of 65 categories with 9,213 images, and the COCO-SEG dataset contains 200k images of 78 categories. The combinations of training datasets primarily used by existing methods include D+C \cite{li2023discriminative}, \cite{zhu2023co}, \cite{zhang2021summarize}, \cite{jin2020icnet}, as well as D+S \cite{wu2023co}, \cite{xu2023co}, \cite{zheng2023memory}. In the comparative experiments, we conduct experiments using both combinations of training data to ensure fairness of comparison. In all ablation experiments, we use the smaller datasets DUT-class and COCO-9k as the training set.
	
	\textbf{Quantitative Results.} Table 1 presents a quantitative comparison of our VCP with the most representative works in the past three years across multiple metrics. Compared to 12 full fine-tuning CoSOD methods, our VCP demonstrates a significant performance advantage on three commonly used benchmark test sets. Particularly on the CoCA dataset, which is the most challenging and best reflects the essence of the CoSOD task, our method outperforms the second-best by 5.6$\%$ and 6.8$\%$ in terms of $S_m$ and $F_m$ metrics, respectively. Compared to the prompt-based tuning method EVP \cite{liu2023explicit}, our VCP exhibits overwhelmingly superior performance. Comprehensive experiments demonstrate the promising prospects of introducing prompt learning for CoSOD tasks and validate the powerful capability of our proposed concise and parameter-efficient VCP in addressing this task. The Precision-Recall and $F_m$-Threshold curves shown in Fig. \textbf{S1} further demonstrate the effectiveness of our method compared to 12 state-of-the-art methods.
	
			\begin{table}[!t]
		\centering
		\normalsize
		\tabcolsep=0.005cm
		\renewcommand{\arraystretch}{1.1}
		\caption*{\textbf{Table S1}: Ablation on the tuning stages, i.e., tuning different Transformer blocks of the foundation model (SegFormer).} 
		\vspace{-0.1in} 
		\scalebox{0.73}{
			\begin{tabular}{c|c|ccc|ccc|ccc}
				\toprule
				\multirow{2}[4]{*}{Tuning Stage} & \multicolumn{1}{p{4.29em}|}{\multirow{2}[2]{*}{\makecell[c]{Tunable \\ Param. (M)}}} & \multicolumn{3}{c|}{COCA} & \multicolumn{3}{c|}{CoSOD3k} & \multicolumn{3}{c}{CoSal2015} \\
				\cmidrule{3-11}          &       & $S_m$↑   & $F_m^{\max}$↑ & MAE↓  & $S_m$↑   & $F_m^{\max}$↑ & MAE↓  & $S_m$↑   & $F_m^{\max}$↑ & MAE↓  \\
				\midrule
				Stage-1 & 1.82  & .683 & .532 & .123 & .841 & .821 & .066 & .883 & .881 & .057 \\
				Stage-1,2 & 2.22  & .699 & .559 & .104 & .847 & .828 & .064 & .887 & .891 & .055 \\
				Stage-1,2,3 & 3.34  & .723 & .594 & .097 & .858 & .836 & .060 & .901 & .903 & .047 \\
				\midrule
				Stage-1,2,3,4 & 4.94  & \textcolor[rgb]{ 1,  0,  0}{\textbf{.774}} & \textcolor[rgb]{ 1,  0,  0}{\textbf{.680}} & \textcolor[rgb]{ 1,  0,  0}{\textbf{.069}} & \textcolor[rgb]{ 1,  0,  0}{\textbf{.874}} & \textcolor[rgb]{ 1,  0,  0}{\textbf{.868}} & \textcolor[rgb]{ 1,  0,  0}{\textbf{.049}} & \textcolor[rgb]{ 1,  0,  0}{\textbf{.911}} & \textcolor[rgb]{ 1,  0,  0}{\textbf{.920}} & \textcolor[rgb]{ 1,  0,  0}{\textbf{.037}} \\
				\bottomrule
			\end{tabular}%
		}
		\vspace{-0.1in}
	\end{table}
	
	\textbf{Qualitative Results.} Fig. 5 illustrates the visual comparisons between our VCP and seven representative works across four selected scenarios. "Normal" denotes scenarios with relatively simple backgrounds and minimal interference from non-co-salient objects. "Distraction" indicates scenarios where there are numerous interfering salient objects apart from the co-salient objects. "Tiny Object" indicates that the co-salient objects in the image have a smaller pixel ratio. "Variation" represents scenarios where co-salient objects within specific groups exhibit various presentation styles. Through the visual comparisons in Fig. 5, it can be observed that our VCP achieves more effective localization and segmentation performance even when facing challenging cases such as multiple salient object interferences, small or diverse co-salient objects. The performance gains are attributed to our task-specific VCP, which adequately stimulates the semantic representations in the foundation model.

	\section{Ablation Analysis on Architecture Configuration and Hyperparameters.}
	
	\textbf{Effectiveness of the tuning stages.} Further experiments are conducted to validate the impact of tuning different stages of the foundation model on the final performance. As shown in Table \textbf{S1}, the improvement in model performance becomes increasingly evident as the prompt addition stage deepens.

	\textbf{Effectiveness of the parameter scale factor $r$.} Table \textbf{S2} presents the ablation analysis of the scale factor used when performing downsampling with MLP on frozen embeddings. The scale factor significantly impacts the tunable parameters of the model, and while $r=8$ strikes a balance between parameters and performance, $r=4$ is ultimately chosen considering the model's performance. 
	\begin{table}[!t]
		\centering
		\normalsize
		\tabcolsep=0.04cm
		\renewcommand{\arraystretch}{1.0}
		\caption*{\textbf{Table S2}: Ablation on the parameter scale factor $r$.} \label{tab-6}
		\vspace{-0.05in} 
		\scalebox{0.73}{
			\begin{tabular}{c|c|ccc|ccc|ccc}
				\toprule
				\multirow{2}[4]{*}{Scale-$r$} & \multicolumn{1}{p{4.29em}|}{\multirow{2}[2]{*}{\makecell[c]{Tunable \\ Param. (M)}}} & \multicolumn{3}{c|}{COCA} & \multicolumn{3}{c|}{CoSOD3k} & \multicolumn{3}{c}{CoSal2015} \\
				\cmidrule{3-11}          &      & $S_m$↑   & $F_m^{\max}$↑ & MAE↓  & $S_m$↑   & $F_m^{\max}$↑ & MAE↓  & $S_m$↑   & $F_m^{\max}$↑ & MAE↓ \\
				\midrule
				$r$=64  & 2.71  & .715 & .582 & .099 & .858 & .842 & .055 & .896 & .901 & .046 \\
				$r$=32  & 2.76  & .740 & .627 & .088 & .865 & .853 & .055 & .905 & .912 & .044 \\
				$r$=16  & 2.89  & .759 & .656 & .081 & .872 & .862 & .053 & .911 & .918 & .039 \\
				$r$=8   & 3.34  & .769 & .665 & \textcolor[rgb]{ 1,  0,  0}{\textbf{.069}} & .872 & .864 & .051 & \textcolor[rgb]{ 1,  0,  0}{\textbf{.913}} & 0.918 & \textcolor[rgb]{ 1,  0,  0}{\textbf{.035}} \\
				\midrule
				$r$=4   & 4.94  & \textcolor[rgb]{ 1,  0,  0}{\textbf{.774}} & \textcolor[rgb]{ 1,  0,  0}{\textbf{.680}} & \textcolor[rgb]{ 1,  0,  0}{\textbf{.069}} & \textcolor[rgb]{ 1,  0,  0}{\textbf{.874}} & \textcolor[rgb]{ 1,  0,  0}{\textbf{.868}} & \textcolor[rgb]{ 1,  0,  0}{\textbf{.049}} & .911 & \textcolor[rgb]{ 1,  0,  0}{\textbf{.920}} & .037 \\
				\bottomrule
			\end{tabular}%
		}
		\vspace{-0.1in}
	\end{table}%
	
	\textbf{Effectiveness of hyperparameters in CPG:}
	The main insight of our CPG is to enforce effective tunable parameters to focus on and mine co-salient representations within intra-group frozen embeddings, thereby generating task-specific consensus prompts ${P_{Co}}$.
	
	\begin{figure}[!t]
		\centering
		\includegraphics[width=3.2in]{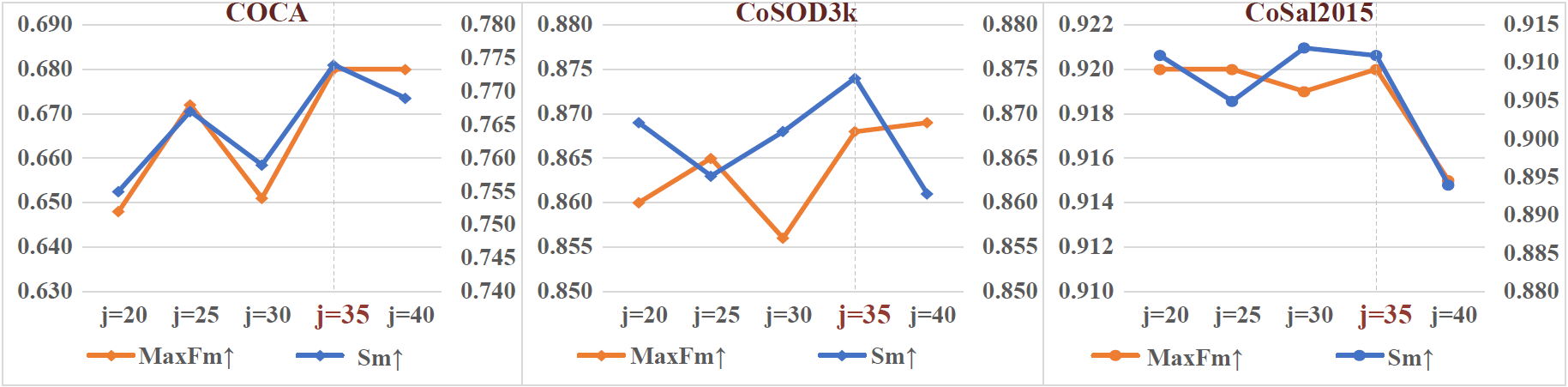}
		\vspace{-0.1in}
		\DeclareGraphicsExtensions.
		\begin{center}
			\vspace{-0.05in}
			\caption*{\textbf{Fig. S2}: Ablation on the predefined Saliency Seeds (SS) quantity $j$ in our CPG. Bold indicates the default settings in our VCP.} \label{fig-7}
		\end{center}
		\vspace{-0.4in}
	\end{figure}
	\begin{table}[!t]
		\centering
		\normalsize
		\tabcolsep=0.036cm
		\renewcommand{\arraystretch}{1.0}
		\caption*{\textbf{Table S3}: Ablation on the number of representative consensus seeds selected in our CPG (top-$k$).} 
		\vspace{-0.05in}
		\scalebox{0.74}{
			\begin{tabular}{c|c|ccc|ccc|ccc}
				\toprule
				\multirow{2}[4]{*}{Top-$k$} & \multicolumn{1}{p{4.29em}|}{\multirow{2}[2]{*}{\makecell[c]{Tunable \\ Param. (M)}}} &
				\multicolumn{3}{c|}{COCA} & \multicolumn{3}{c|}{CoSOD3k} & \multicolumn{3}{c}{CoSal2015} \\
				\cmidrule{3-11}          &       & $S_m$↑   & $F_m^{\max}$↑ & MAE↓  & $S_m$↑   & $F_m^{\max}$↑ & MAE↓  & $S_m$↑   & $F_m^{\max}$↑ & MAE↓ \\
				\midrule
				$k$=16  & 4.94  & .760 & .659 & .085 & .873 & .867 & .055 & .908 & .916 & .046 \\
				$k$=24  & 4.94  & .768 & .669 & .075 & .871 & .862 & .055 & \textcolor[rgb]{ 1,  0,  0}{\textbf{.912}} & \textcolor[rgb]{ 1,  0,  0}{\textbf{.922}} & .041 \\
				\textbf{$k$=32} & 4.94  & \textcolor[rgb]{ 1,  0,  0}{\textbf{.774}} & \textcolor[rgb]{ 1,  0,  0}{\textbf{.680}} & \textcolor[rgb]{ 1,  0,  0}{\textbf{.069}} & .874 & .868 & \textcolor[rgb]{ 1,  0,  0}{\textbf{.049}} & .911 & .920 & \textcolor[rgb]{ 1,  0,  0}{\textbf{.037}} \\
				$k$=40  & 4.95  & .752 & .655 & .079 & \textcolor[rgb]{ 1,  0,  0}{\textbf{.876}} & \textcolor[rgb]{ 1,  0,  0}{\textbf{.870}} & .052 & .906 & .917 & .044 \\
				$k$=48  & 4.96  & .754 & .644 & .076 & .869 & .861 & .052 & .907 & .916 & .040 \\
				\bottomrule
			\end{tabular}%
		}
		\vspace{-0.1in}
	\end{table}

\begin{table}[!t]
	\centering
	\normalsize
	\tabcolsep=0.016cm
	\renewcommand{\arraystretch}{1.1}
	\caption*{\textbf{Table S4}: Ablation experiments are conducted based on different unified mapping dimensions $d$ of the Segformer prediction head. "OOM" indicates out-of-memory.} 
	\vspace{-0.05in}
	\scalebox{0.75}{
		\begin{tabular}{c|c|ccc|ccc|ccc}
			\toprule
			\multirow{2}[2]{*}{\makecell[c]{Segformer \\ Head}} & \multicolumn{1}{p{4.29em}|}{\multirow{2}[2]{*}{\makecell[c]{Tunable \\ Param. (M)}}} & \multicolumn{3}{c|}{COCA} & \multicolumn{3}{c|}{CoSOD3k} & \multicolumn{3}{c}{CoSal2015} \\
			\cmidrule{3-11}          &       & $S_m$↑   & $F_m^{\max}$↑ & MAE↓  & $S_m$↑   & $F_m^{\max}$↑ & MAE↓  & $S_m$↑   & $F_m^{\max}$↑ & MAE↓ \\
			\midrule
			\textbf{$d=$768} & 7.83  & \multicolumn{3}{c|}{OOM} & \multicolumn{3}{c|}{OOM} & \multicolumn{3}{c}{OOM} \\
			$d=$384 & 4.98  & \multicolumn{3}{c|}{OOM} & \multicolumn{3}{c|}{OOM} & \multicolumn{3}{c}{OOM} \\
			$d=$128 & 3.84  & .747 & .656 & .101 & .872 & .861 & .054 & .909 & .916 & .038 \\
			\bottomrule
		\end{tabular}%
	}
	\vspace{-0.1in}
\end{table}%
\begin{table}[!t]
	\centering
	\normalsize
	\tabcolsep=0.038cm
	\renewcommand{\arraystretch}{1.12}
	\caption*{\textbf{Table S5}: Ablation regarding the preset parameter $d$ in PH.} 
	\vspace{-0.05in}
	\scalebox{0.75}{
		\begin{tabular}{c|c|ccc|ccc|ccc}
			\toprule
			\multirow{2}[2]{*}{PH-$d$} & \multicolumn{1}{p{4.29em}|}{\multirow{2}[2]{*}{\makecell[c]{Tunable \\ Param. (M)}}} &
			\multicolumn{3}{c|}{COCA} & \multicolumn{3}{c|}{CoSOD3k} & \multicolumn{3}{c}{CoSal2015} \\
			\cmidrule{3-11}          &       & $S_m$↑   & $F_m^{\max}$↑ & MAE↓  & $S_m$↑   & $F_m^{\max}$↑ & MAE↓  & $S_m$↑   & $F_m^{\max}$↑ & MAE↓ \\
			\midrule
			$d$=64  & 4.53  & .762 & .658 & .075 & .870 & .862 & .051 & .907 & .914 & .040 \\
			$d$=96  & 4.74  & .766 & .661 & .080 & .871 & .865 & .058 & .909 & \textcolor[rgb]{ 1,  0,  0}{\textbf{.921}} & .051 \\
			\textbf{$d$=128} & 4.94  & \textcolor[rgb]{ 1,  0,  0}{\textbf{.774}} & \textcolor[rgb]{ 1,  0,  0}{\textbf{.680}} & \textcolor[rgb]{ 1,  0,  0}{\textbf{.069}} & .874 & \textcolor[rgb]{ 1,  0,  0}{\textbf{.868}} & .049 & \textcolor[rgb]{ 1,  0,  0}{\textbf{.911}} & .920 & \textcolor[rgb]{ 1,  0,  0}{\textbf{.037}} \\
			$d$=192 & 5.39  & .757 & .647 & .082 & \textcolor[rgb]{ 1,  0,  0}{\textbf{.878}} & .867 & \textcolor[rgb]{ 1,  0,  0}{\textbf{.048}} & .910 & .913 & \textcolor[rgb]{ 1,  0,  0}{\textbf{.037}} \\
			$d$=256 & 5.87  & .749 & .633 & .084 & .871 & .852 & .051 & .909 & .912 & \textcolor[rgb]{ 1,  0,  0}{\textbf{.037}} \\
			\bottomrule
		\end{tabular}%
	}
	\vspace{-0.1in}
\end{table}

	\begin{table*}[!t]
	\centering
	\normalsize
	\tabcolsep=0.14cm
	\renewcommand{\arraystretch}{1.0}
	\caption*{\textbf{Table S6}: Some alternative schemes of VCP exhibit more efficient parameters and competitive performance. $r$ represents the parameter scale factor  (default $r=4$), $d$ represents the unified mapping dimension in PH  (default $d=128$), and "Share MLP" indicates the adoption of stage-shared MLP in CPD (default as adaptive tuning).} 
	\vspace{-0.05in}
	\scalebox{0.85}{
		\begin{tabular}{c|c|c|c|ccc|ccc|ccc}
			\toprule
			\multicolumn{1}{c|}{\multirow{2}[4]{*}{Combination}} & 
			\multicolumn{1}{c|}{\multirow{2}[2]{*}{\makecell[c]{Train \\ Dataset}}} & \multicolumn{1}{c|}{\multirow{2}[2]{*}{\makecell[c]{Tunable \\ Param. (M)}}} & 			\multicolumn{1}{c|}{\multirow{2}[2]{*}{\makecell[c]{Model Size \\ (MB)}}} & \multicolumn{3}{c|}{COCA} & \multicolumn{3}{c|}{CoSOD3k} & \multicolumn{3}{c}{CoSal2015} \\
			\cmidrule{5-13}          &       &       &       & $S_m$↑   & $F_m^{\max}$↑ & MAE↓  & $S_m$↑   & $F_m^{\max}$↑ & MAE↓  & $S_m$↑   & $F_m^{\max}$↑ & MAE↓ \\
			\midrule
			\multicolumn{1}{c|}{EVP} & S+D   & 3.70  & 14.10 & 0.686 & 0.546 & 0.126 & 0.839 & 0.813 & 0.076 & 0.876 & 0.874 & 0.068 \\
			\midrule
			VCP($r=8+d=96$) & S+D   & 3.13  & 12.08 & 0.814 & 0.745 & 0.054 & \textcolor[rgb]{ 1,  0,  0}{\textbf{0.895}} & 0.892 & \textcolor[rgb]{ 1,  0,  0}{\textbf{0.043}} & 0.922 & 0.933 & 0.034 \\
			VCP($r=8+$Share MLP) & S+D   & 3.24  & 12.47 & 0.817 & \textcolor[rgb]{ 1,  0,  0}{\textbf{0.753}} & \textcolor[rgb]{ 1,  0,  0}{\textbf{0.053}} & 0.889 & 0.887 & 0.047 & 0.918 & 0.926 & 0.035 \\
			\textbf{VCP($r=4$)} & S+D   & 4.94  & 19.02  & \textcolor[rgb]{ 1,  0,  0}{\textbf{0.819}} & 0.752 & 0.054 & \textcolor[rgb]{ 1,  0,  0}{\textbf{0.895}} & \textcolor[rgb]{ 1,  0,  0}{\textbf{0.893}} & \textcolor[rgb]{ 1,  0,  0}{\textbf{0.043}} & \textcolor[rgb]{ 1,  0,  0}{\textbf{0.927}} & \textcolor[rgb]{ 1,  0,  0}{\textbf{0.941}} & \textcolor[rgb]{ 1,  0,  0}{\textbf{0.030}} \\
			\midrule
			VCP($r=8+d=96$) & C+D   & 3.13  & 12.08 & 0.770 & 0.672 & 0.079 & 0.873 & \textcolor[rgb]{ 0,  0,  1}{0.869} & 0.050 & 0.911 & 0.919 & 0.038 \\
			VCP($r=8+$Share MLP) & C+D   & 3.24  & 12.47 & 0.767 & 0.665 & 0.072 & 0.870 & 0.862 & 0.051 & 0.909 & 0.918 & 0.040 \\
			VCP($r=8$) & C+D   & 3.34  & 12.89 & 0.769 & 0.665 & \textcolor[rgb]{ 0,  0,  1}{0.069} & 0.872 & 0.864 & 0.051 & \textcolor[rgb]{ 0,  0,  1}{0.913} & 0.918 & \textcolor[rgb]{ 0,  0,  1}{0.035} \\
			\textbf{VCP($r=4$)} & C+D   & 4.94  & 19.02  & \textcolor[rgb]{ 0,  0,  1}{0.774} & \textcolor[rgb]{ 0,  0,  1}{0.680} & \textcolor[rgb]{ 0,  0,  1}{0.069} & \textcolor[rgb]{ 0,  0,  1}{0.874} & 0.868 & \textcolor[rgb]{ 0,  0,  1}{0.049} & 0.911 & \textcolor[rgb]{ 0,  0,  1}{0.920} & 0.037 \\
			\bottomrule
		\end{tabular}%
	}
	\vspace{-0.1in}
\end{table*}%
	We achieve our CPG by predefining $j$ learnable saliency seeds, clustering representative prototypes of salient objects to generate saliency estimation maps $\{ {M^s}\} _{s = 1}^4$ for each individual images. As shown in Fig. \textbf{S2}, different numbers of saliency seeds result in certain fluctuations in the model's tunable parameters, but all ensure competitive performance gains. Considering both the effectiveness of performance and the efficiency of tunable parameters, $j=35$ is selected.
	
	The generated saliency estimation maps are utilized to focus on pixel embeddings to form consensus seeds $C{o_{seed}} \in {\mathbb{R}^{N{L_s} \times {C_r}}}$. Finally, we select the top-$k$ most relevant consensus seeds to form representative consensus $Co_{seed}^{rep} \in {\mathbb{R}^{k \times {C_r}}}$. We further conduct experiments to validate the effectiveness of the predefined saliency seeds and the number of selected representative consensus seeds (top-$k$). As shown in Table \textbf{S3}, the quantity of representative consensus seeds has a minimal impact on the overall number of tunable parameters in the model, but it does introduce some performance fluctuations. For the comprehensive effectiveness of the model's performance, $k=32$ is determined as the default setting.
	
	\textbf{Effectiveness of hyperparameters in PH:} As shown in Table \textbf{S4}, segformer's original prediction head up-projects multi-scale features onto a unified high dimensionality ($d$=768) and employs multiple fully connected layers for prediction, resulting in a higher number of tunable parameters (3.15M).
	
	Hence, a more concise prediction head (PH) is designed (1.49M). Our PH initially projects multi-scale features to a lower dimensionality ($d=128$). Subsequently, the deepest features are processed with ASPP and utilized to guide FPN-like decoding and input to a linear classifier for category prediction. We conducted ablation experiments on the set unified dimensionality $d$. As shown in Table \textbf{S5}, the hyperparameter $d$ is not the key factor affecting the overall parameter count of the model. It can be observed that $d=96$ demonstrates certain performance competitiveness while reducing a certain number of tunable parameters. Considering both the effectiveness of performance and the efficiency of tunable parameters, $d=128$ is selected.
	
	\begin{table}[!t]
		\centering
		\normalsize
		\tabcolsep=0.0001cm
		\renewcommand{\arraystretch}{1.0}
		\caption*{\textbf{Table S7}: Comprehensive comparison with 5 cutting-edge methods and ablation analysis of the used backbone.}
		\vspace{-0.05in}
		\scalebox{0.66}{
			\begin{tabular}{cc|cccccccc}
				\toprule
				\multicolumn{1}{c|}{\multirow{2}[2]{*}{Datasets}} & \multicolumn{1}{c|}{\multirow{2}[2]{*}{Metrics}} & CoRP  & \multicolumn{1}{c}{\textbf{OURS$_I$}} & \multicolumn{1}{c}{EVP} & \multicolumn{1}{c}{GEM} & \multicolumn{1}{c}{MCCL} & \multicolumn{1}{c}{SCED} & \multicolumn{1}{c}{\textbf{OURS$_{II}$}} & \multicolumn{1}{c}{\textbf{OURS$_{II}$}}  \\
				\multicolumn{1}{c|}{} &       & TPAMI$_{23}$ & \multicolumn{1}{c}{—} & \multicolumn{1}{c}{CVPR$_{23}$} & \multicolumn{1}{c}{CVPR$_{23}$} & \multicolumn{1}{c}{AAAI$_{23}$} & \multicolumn{1}{c}{ACMM$_{23}$} & \multicolumn{1}{c}{—} & \multicolumn{1}{c}{—} \\
				\midrule
				\multicolumn{1}{c|}{\multirow{4}[2]{*}{CoCA}} & \multicolumn{1}{c|}{$S_m$↑} & \multicolumn{1}{c}{0.732} & \textcolor[rgb]{ 0,  0,  1}{0.774} & 0.686 & 0.726 & 0.713 & 0.741 & \textcolor[rgb]{ 1,  0,  0}{\textbf{0.819}} & 0.805  \\
				\multicolumn{1}{c|}{} & \multicolumn{1}{c|}{$F_m^{\max}$↑} & \multicolumn{1}{c}{0.619} & \textcolor[rgb]{ 0,  0,  1}{0.680} & 0.546 & 0.599 & 0.584 & 0.629 & \textcolor[rgb]{ 1,  0,  0}{\textbf{0.752}} & 0.731 \\
				\multicolumn{1}{c|}{} & \multicolumn{1}{c|}{$MAE$↓} & \multicolumn{1}{c}{0.093} & \textcolor[rgb]{ 0,  0,  1}{0.069} & 0.126 & 0.095 & 0.097 & 0.084 & \textcolor[rgb]{ 1,  0,  0}{\textbf{0.054}} & 0.059  \\
				\multicolumn{1}{c|}{} & \multicolumn{1}{c|}{$E_m$↑} & \multicolumn{1}{c}{0.773} & \textcolor[rgb]{ 0,  0,  1}{0.813} & 0.708 & 0.767 & 0.764 & 0.804 & \textcolor[rgb]{ 1,  0,  0}{\textbf{0.830}} & 0.823  \\
				\midrule
				\multicolumn{2}{l|}{{\small Training Dataset}} & C+D   & \multicolumn{1}{c}{C+D} & \multicolumn{1}{c}{S+D} & \multicolumn{1}{c}{S+D} & \multicolumn{1}{c}{S+D} & \multicolumn{1}{c}{S+D} & \multicolumn{1}{c}{S+D} & \multicolumn{1}{c}{S+D}  \\
				\midrule
				\multicolumn{2}{l|}{{\small Foundation Model}} & {\small PVTv2} & \multicolumn{1}{c}{{\small SegF}} & \multicolumn{1}{c}{{\small SegF}} & \multicolumn{1}{c}{{\small T2T-ViT}} & \multicolumn{1}{c}{PVTv2} & \multicolumn{1}{c}{{\small CLIP-ViT}} & \multicolumn{1}{c}{{\small SegF}} & \multicolumn{1}{c}{{\small PVTv2}} \\
				\midrule
				\multicolumn{2}{l|}{{\small Tunable Param. (M)}} & —     & 4.9   & 3.7   & 52.3  & 27.1  & 156.7  & 4.9  & 4.95   \\
				\midrule
				\multicolumn{2}{l|}{{\small Model Size (MB)}} & —     & 19    & 14.1  & 199.7 & 104.5 & 1750  & 19  & 19 \\
				\bottomrule
			\end{tabular}%
		}
		\vspace{-0.1in}
	\end{table}%

\begin{table*}[!t]
	\centering
	\normalsize
	\tabcolsep=0.07cm
	\renewcommand{\arraystretch}{1.0}
	\caption*{\textbf{Table S8}: Generalization verification on the RGB-D CoSOD task. Our VCP is retrained with the same training set (DUT-class \cite{zhang2020gradient}) as existing RGB-D CoSOD methods and conducts testing on three benchmark datasets (CoSal1k \cite{zhang2022learning}, CoSal150 \cite{zhang2016detection} and CoSal183 \cite{fu2015object}).} 
	\vspace{-0.06in}
	\scalebox{0.83}{
		\begin{tabular}{cc|ccccccccccccc}
			\toprule
			\multicolumn{1}{c|}{\multirow{2}[2]{*}{Datasets}} & \multicolumn{1}{c|}{\multirow{2}[2]{*}{Metrics}} & HSCS \cite{cong2018hscs} & CBCS \cite{fu2013cluster} & \multicolumn{1}{c}{CoEG \cite{fan2021re}} & \multicolumn{1}{c}{GICD} & \multicolumn{1}{c}{GICD$_{+D}$} & \multicolumn{1}{c}{ICNet} & \multicolumn{1}{c}{ICNet$_{+D}$} & \multicolumn{1}{c}{GCoN} & \multicolumn{1}{c}{GCoN$_{+D}$} & \multicolumn{1}{c}{CADC} & \multicolumn{1}{c}{CADC$_{+D}$} & \multicolumn{1}{c}{CTNet} & \multicolumn{1}{c}{\multirow{2}[2]{*}{OURS}} \\
			\multicolumn{1}{c|}{} &       & TMM$_{19}$ & TIP$_{22}$ & \multicolumn{1}{c}{TPAMI$_{21}$} & \multicolumn{2}{c}{ECCV$_{20}$ \cite{zhang2020gradient}} & \multicolumn{2}{c}{NeurIPS$_{20}$ \cite{jin2020icnet}} & \multicolumn{2}{c}{CVPR$_{21}$ \cite{fan2021group}} & \multicolumn{2}{c}{ICCV$_{21}$ \cite{zhang2021summarize}}  & \multicolumn{1}{c}{TIP$_{22}$ \cite{zhang2022learning}} &  \\
			\midrule
			\multicolumn{1}{c|}{\multirow{6}[2]{*}{CoSal1k}} & \multicolumn{1}{c|}{$S_m$↑} & \multicolumn{1}{c}{0.478} & \multicolumn{1}{c}{0.529} & 0.811 & 0.793 & 0.756 & 0.840 & 0.788 & 0.807 & 0.810 & 0.850 & 0.841 & \textcolor[rgb]{ 0,  0,  1}{0.875} & \textcolor[rgb]{ 1,  0,  0}{\textbf{0.893}} \\
			\multicolumn{1}{c|}{} & \multicolumn{1}{c|}{$E_m^{\max}$↑} & \multicolumn{1}{c}{0.640} & \multicolumn{1}{c}{0.651} & 0.866 & 0.846 & 0.816 & 0.895 & 0.844 & 0.864 & 0.861 & 0.892 & 0.885 & \textcolor[rgb]{ 0,  0,  1}{0.913} & \textcolor[rgb]{ 1,  0,  0}{\textbf{0.933}} \\
			\multicolumn{1}{c|}{} & \multicolumn{1}{c|}{$F_m^{\max}$↑} & \multicolumn{1}{c}{0.428} & \multicolumn{1}{c}{0.505} & 0.799 & 0.783 & 0.746 & 0.846 & 0.767 & 0.814 & 0.815 & 0.837 & 0.832 & \textcolor[rgb]{ 0,  0,  1}{0.865} & \textcolor[rgb]{ 1,  0,  0}{\textbf{0.896}} \\
			\multicolumn{1}{c|}{} & \multicolumn{1}{c|}{MAE↓} & \multicolumn{1}{c}{0.242} & \multicolumn{1}{c}{0.230} & 0.085 & 0.090 & 0.102 & 0.073 & 0.093 & 0.083 & 0.080 & 0.077 & 0.090 & \textcolor[rgb]{ 0,  0,  1}{0.063} & \textcolor[rgb]{ 1,  0,  0}{\textbf{0.045}} \\
			\multicolumn{1}{c|}{} & \multicolumn{1}{c|}{$E_m$↑} & \multicolumn{1}{c}{0.512} & \multicolumn{1}{c}{0.520} & 0.815 & 0.842 & 0.809 & 0.891 & 0.839 & 0.856 & 0.855 & 0.861 & 0.835 & \textcolor[rgb]{ 0,  0,  1}{0.882} & \textcolor[rgb]{ 1,  0,  0}{\textbf{0.921}} \\
			\multicolumn{1}{c|}{} & \multicolumn{1}{c|}{$F_m$↑} & \multicolumn{1}{c}{0.293} & \multicolumn{1}{c}{0.354} & 0.770 & 0.778 & 0.743 & 0.836 & 0.754 & 0.809 & 0.810 & 0.802 & 0.779 & \textcolor[rgb]{ 0,  0,  1}{0.834} & \textcolor[rgb]{ 1,  0,  0}{\textbf{0.881}} \\
			\midrule
			\multicolumn{1}{c|}{\multirow{6}[2]{*}{CoSal150}} & \multicolumn{1}{c|}{$S_m$↑} & \multicolumn{1}{c}{0.661} & \multicolumn{1}{c}{0.576} & 0.851 & 0.821 & 0.812 & 0.888 & 0.856 & 0.821 & 0.867 & 0.895 & 0.895 & \textcolor[rgb]{ 0,  0,  1}{0.910} & \textcolor[rgb]{ 1,  0,  0}{\textbf{0.920}} \\
			\multicolumn{1}{c|}{} & \multicolumn{1}{c|}{$E_m^{\max}$↑} & \multicolumn{1}{c}{0.884} & \multicolumn{1}{c}{0.726} & 0.890 & 0.878 & 0.880 & 0.939 & 0.920 & 0.876 & 0.924 & 0.933 & 0.759 & \textcolor[rgb]{ 0,  0,  1}{0.944} & \textcolor[rgb]{ 1,  0,  0}{\textbf{0.953}} \\
			\multicolumn{1}{c|}{} & \multicolumn{1}{c|}{$F_m^{\max}$↑} & \multicolumn{1}{c}{0.836} & \multicolumn{1}{c}{0.608} & 0.870 & 0.846 & 0.831 & 0.906 & 0.867 & 0.834 & 0.878 & 0.893 & 0.905 & \textcolor[rgb]{ 0,  0,  1}{0.912} & \textcolor[rgb]{ 1,  0,  0}{\textbf{0.924}} \\
			\multicolumn{1}{c|}{} & \multicolumn{1}{c|}{MAE↓} & \multicolumn{1}{c}{0.163} & \multicolumn{1}{c}{0.211} & 0.079 & 0.077 & 0.089 & 0.050 & 0.069 & 0.080 & 0.059 & 0.062 & 0.068 & \textcolor[rgb]{ 0,  0,  1}{0.053} & \textcolor[rgb]{ 1,  0,  0}{\textbf{0.039}} \\
			\multicolumn{1}{c|}{} & \multicolumn{1}{c|}{$E_m$↑} & \multicolumn{1}{c}{0.605} & \multicolumn{1}{c}{0.550} & 0.873 & 0.874 & 0.876 & 0.935 & 0.910 & 0.871 & 0.921 & 0.904 & 0.892 & \textcolor[rgb]{ 0,  0,  1}{0.920} & \textcolor[rgb]{ 1,  0,  0}{\textbf{0.944}} \\
			\multicolumn{1}{c|}{} & \multicolumn{1}{c|}{$F_m$↑} & \multicolumn{1}{c}{0.590} & \multicolumn{1}{c}{0.443} & 0.850 & 0.840 & 0.876 & 0.896 & 0.844 & 0.828 & 0.869 & 0.858 & 0.848 & \textcolor[rgb]{ 0,  0,  1}{0.880} & \textcolor[rgb]{ 1,  0,  0}{\textbf{0.905}} \\
			\midrule
			\multicolumn{1}{c|}{\multirow{6}[2]{*}{CoSal183}} & \multicolumn{1}{c|}{$S_m$↑} & \multicolumn{1}{c}{0.704} & \multicolumn{1}{c}{0.620} & 0.728 & 0.645 & 0.661 & 0.763 & 0.719 & 0.707 & 0.708 & 0.709 & 0.649 & \textcolor[rgb]{ 0,  0,  1}{0.764} & \textcolor[rgb]{ 1,  0,  0}{\textbf{0.795}} \\
			\multicolumn{1}{c|}{} & \multicolumn{1}{c|}{$E_m^{\max}$↑} & \multicolumn{1}{c}{0.790} & \multicolumn{1}{c}{0.748} & 0.764 & 0.668 & 0.687 & 0.768 & 0.757 & 0.726 & 0.719 & 0.770 & 0.759 & \textcolor[rgb]{ 0,  0,  1}{0.837} & \textcolor[rgb]{ 1,  0,  0}{\textbf{0.873}} \\
			\multicolumn{1}{c|}{} & \multicolumn{1}{c|}{$F_m^{\max}$↑} & \multicolumn{1}{c}{0.603} & \multicolumn{1}{c}{0.486} & 0.558 & 0.442 & 0.448 & 0.617 & 0.548 & 0.553 & 0.556 & 0.609 & 0.592 & \textcolor[rgb]{ 0,  0,  1}{0.654} & \textcolor[rgb]{ 1,  0,  0}{\textbf{0.699}} \\
			\multicolumn{1}{c|}{} & \multicolumn{1}{c|}{MAE↓} & \multicolumn{1}{c}{0.079} & \multicolumn{1}{c}{0.088} & 0.073 & 0.111 & 0.100 & 0.066 & 0.073 & 0.088 & 0.094 & 0.110 & 0.121 & \textcolor[rgb]{ 0,  0,  1}{0.068} & \textcolor[rgb]{ 1,  0,  0}{\textbf{0.051}} \\
			\multicolumn{1}{c|}{} & \multicolumn{1}{c|}{$E_m$↑} & \multicolumn{1}{c}{0.710} & \multicolumn{1}{c}{0.636} & 0.742 & 0.655 & 0.673 & 0.751 & 0.742 & 0.703 & 0.705 & 0.680 & 0.668 & \textcolor[rgb]{ 0,  0,  1}{0.752} & \textcolor[rgb]{ 1,  0,  0}{\textbf{0.790}} \\
			\multicolumn{1}{c|}{} & \multicolumn{1}{c|}{$F_m$↑} & \multicolumn{1}{c}{0.529} & \multicolumn{1}{c}{0.399} & 0.535 & 0.433 & 0.438 & 0.600 & 0.533 & 0.531 & 0.537 & 0.524 & 0.505 & \textcolor[rgb]{ 0,  0,  1}{0.583} & \textcolor[rgb]{ 1,  0,  0}{\textbf{0.630}} \\
			\midrule
			\multicolumn{2}{c|}{Model Size (MB)} & —     & —     & 69    & 533   & 591   & 70    & 163   & 542   & 600   & 1499  & 1596  & 570   & \textbf{19.02} \\
			\midrule
			\multicolumn{2}{c|}{Running Time (FPS)} & 0.12     & 1     & 0.43    & 55.56   & 40   & 76.9    & 71.4   & 62.5   & 41.32   & 31.25  & 19.23  & 17.54   & \textbf{40.4} \\
			\bottomrule
		\end{tabular}
	}
\end{table*}%
\begin{figure*}[!t]
	\centering
	\includegraphics[width=6.8in]{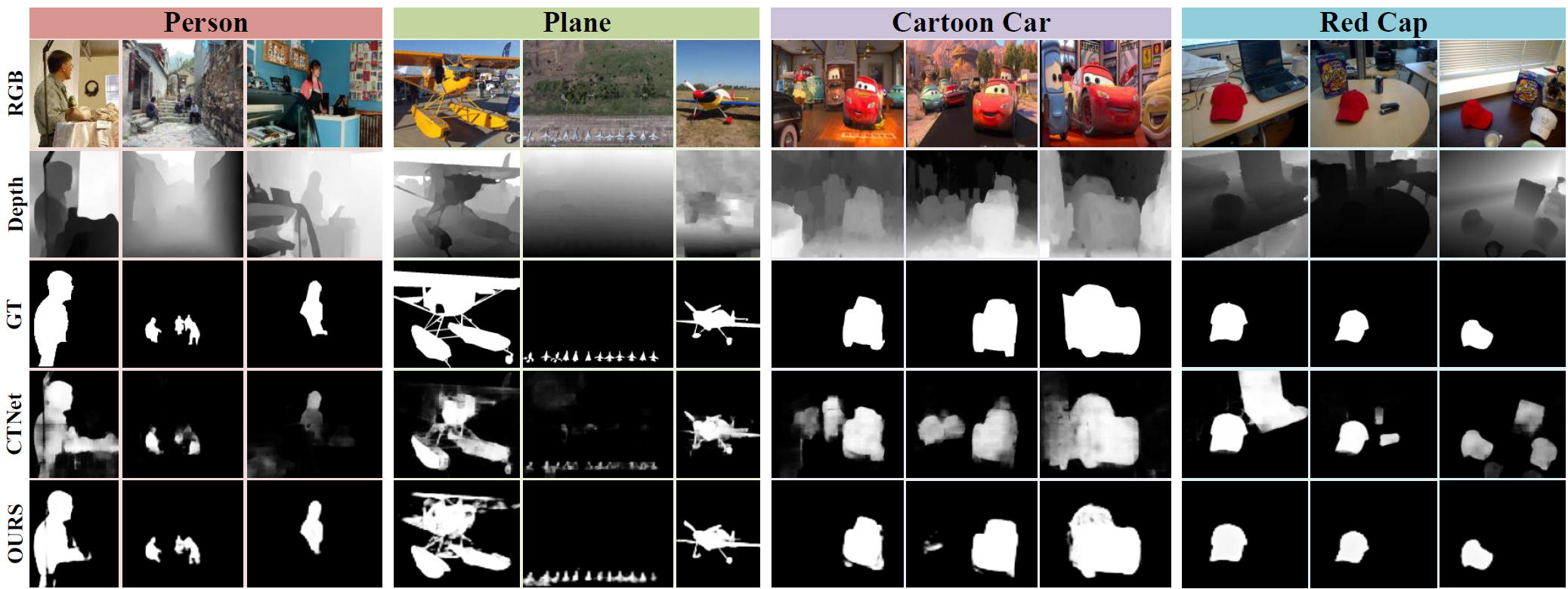}
	\vspace{-0.2in}
	\DeclareGraphicsExtensions.
	\begin{center}
		\caption*{\textbf{Fig. S3}: Visual comparison between our VCP and the most representative RGB-D CoSOD method CTNet \cite{zhang2022learning}.}
	\end{center}
	\vspace{-0.35in}
\end{figure*}

	\textbf{Comparison and analysis regarding the foundation models used.} For fairness in comparison, we use foundation models of comparable scale to existing methods (Table \textbf{S7}). Additionally, without optimizing hyperparameters, we conduct ablation experiments on an additional baseline model, PVTv2 \cite{wang2022pvt}. Through comparison, it is observed that using foundation models of similar or even identical scale, our proposed parameter-efficient method achieves a substantial performance improvement.
	
	\section{Parameter-efficient combination schemes}
	We conduct additional experiments to validate the potential of parameter efficiency while ensuring competitive performance gains. Through extensive analysis of ablation experiments, we find that three main predefined parameters significantly impact the overall parameter scale: the parameter scale factor $r$ in CPG, the unified mapping dimension $d$ in PH, and the fine-tuning mode in CPD. Based on the performance from the conducted ablation experiments as a reference, and considering the competitive scale of tunable parameters compared to existing prompt tuning methods (3.7M), some combinations of VCP schemes are implemented. 
	
	As shown in Table \textbf{S6}, our supplementary three lightweight schemes (with a minimum of only 3.13M tunable parameters) exhibit significantly better parameter efficiency and overwhelming performance effectiveness compared to the state-of-the-art method EVP (with 3.7M tunable parameters). These lightweight schemes achieve competitive performance while maintaining a reduced parameter footprint, making them strong alternatives to our default settings.

	\section{Generalization verification on the RGB-D CoSOD}
	To validate the high generalization capability of the proposed VCP for CoSOD-related tasks, we transfer VCP to address the bimodal RGB-D CoSOD task. For fairness of comparison, we retrain VCP with the same training data as existing methods and conduct tests on three benchmark datasets. Specifically, DUT-class (consists of
	291 categories with 8,250 images) \cite{zhang2020gradient} is used as the training set, and corresponding depth data is generated using depth estimation algorithms \cite{ranftl2021vision}. The three widely used RGB-D CoSOD test sets include: CoSal1k \cite{zhang2022learning} (with 106 groups, totaling 1000 image pairs), CoSal150 \cite{zhang2016detection} (with 21 groups, totaling 150 image pairs), and CoSal183 \cite{fu2015object} (with 16 groups, totaling 183 image pairs).

	To provide more convincing evidence of the effectiveness of the proposed VCP on this task, we refrain from performing any model fine-tuning or post-processing. Furthermore, unlike existing methods that employ dual-stream architectures and specialized cross-modal fusion modules to enhance performance, we utilize a simple single-stream architecture and employ the most basic early fusion strategy by simply adding the cross-modal images as inputs. Comprehensive experiments demonstrate that our simplified version of VCP still achieves remarkable performance on RGB-D CoSOD tasks. As shown in Fig. \textbf{S3}, in scenarios with some interference and insufficient depth effectiveness, our VCP still achieves superior segmentation performance. Table \textbf{S8} presents the quantitative comparison results with state-of-the-art RGB-D CoSOD methods on three benchmark datasets, providing strong evidence of the high generalization capability of our VCP for related group-based segmentation tasks.
\clearpage
{
    \small
    \bibliographystyle{ieeenat_fullname}
    \bibliography{main}
}